\documentclass[sigconf]{acmart}

\renewcommand\footnotetextcopyrightpermission[1]{} % removes footnote with conference information in first column
\usepackage{amsmath,amssymb,amsfonts}
\usepackage{algorithm, algorithmic}
\usepackage{graphicx}
\usepackage{textcomp}
\usepackage{balance}

\usepackage[caption=false, font=footnotesize]{subfig}

\usepackage{booktabs,multirow,tabularx}
\newcommand{\makecell}[2][@{}c@{}]{\begin{tabular}{#1}#2\end{tabular}}

\def\BibTeX{{\rm B\kern-.05em{\sc i\kern-.025em b}\kern-.08emT\kern-.1667em\lower.7ex\hbox{E}\kern-.125emX}}

\begin{document}
\fancyhead{}

\title{Unsupervised Single-Image Reflection Separation \\ Using Perceptual Deep Image Priors}

\author{Suhong Kim}

\affiliation{%
  \institution{School of Computing Science}
  \institution{Simon Fraser University}
  \city{Vancouver}
  \state{British Columbia}
  \postcode{V5A 1S6}
}
\email{suhong_kim@sfu.ca}

\author{Hamed RahmaniKhezri}
\affiliation{%
  \institution{School of Computing Science}
  \institution{Simon Fraser University}
  \city{Vancouver}
  \state{British Columbia}
  \postcode{V5A 1S6}
}
\email{hamed_rahmani@sfu.ca}

\author{Seyed Mohammad Nourbakhsh}
% \authornotemark[2]
\affiliation{%
  \institution{School of Computing Science}
  \institution{Simon Fraser University}
  \city{Vancouver}
  \state{British Columbia}
  \postcode{V5A 1S6}
}
\email{seyed_mohammad_nourbakhsh@sfu.ca}

\author{Mohamed Hefeeda}
% \authornotemark[3]
\affiliation{%
  \institution{School of Computing Science}
  \institution{Simon Fraser University}
  \city{Vancouver}
  \state{British Columbia}
  \postcode{V5A 1S6}
}
\email{mhefeeda@sfu.ca}

\begin{abstract}
Reflections often degrade the quality of the image by obstructing the background scene. This is not desirable for everyday users, and it negatively impacts the performance of multimedia applications that process images with reflections. Most current methods for removing reflections utilize supervised-learning models. However, these models require an extensive number of image pairs to perform well, especially on natural images with reflection, which is difficult to achieve in practice.  In this paper, we propose a novel unsupervised framework for single-image reflection separation. Instead of learning from a large dataset, we optimize the parameters of two cross-coupled deep convolutional networks on a target image to generate two exclusive background and reflection layers. In particular, we design a new architecture of the network to embed semantic features extracted from a pre-trained deep classification network, which gives more meaningful separation similar to human perception. Quantitative and qualitative  results on commonly used datasets in the literature show that our method's performance is at least on par with the state-of-the-art supervised methods and, occasionally, better without requiring large training datasets. Our results also show that our method significantly outperforms the closest unsupervised method in the literature for removing reflections from single images.      
\end{abstract}


\maketitle

\section{Introduction} \label{sec:introduction}

We frequently encounter unpleasant reflections obstructing the scene when taking photos through a transparent surface such as a glass window.  These reflections reduce the visual quality and utility of the images. Reflections may also significantly degrade the performance of multimedia applications such as object detection and face identification. Thus, removing reflection from images is an important problem for users and applications. 

Removing reflections from a single image, however, is not a trivial problem since we need to recover two unknown scenes from a single observation. Specifically, the corrupted image (i.e., the image containing reflections) \textbf{$I$} can be defined as a linear superposition of two image layers: background layer \textbf{$B$} and reflection layer as:  \textbf{$R$}, 
\begin{align}
 I = B + R.  
 \label{eq1}
\end{align}
This expression implies that the problem is inherently ill-posed, since the valid decomposition pairs of \textbf{$B$} and \textbf{$R$} are infinite. 

To address the difficulty of the problem, some prior approaches utilized additional information such as motion cues from a sequence of images captured for the same scene \cite{xue2015computational, sun2016automatic, Han_2017_CVPR, nandoriya2017video, alayrac2019visual, fan2019deep}. In many practical scenarios, a sequence of images of the same scene is not available, and thus these methods would fail. Other prior approaches make a particular assumption on the background and reflection layers, such as the sparse gradient prior \cite{levin2004separating, levin2007user}, the blurriness of the reflection layer \cite{li2014single}, and the ghosting cues \cite{shih2015reflection}. These approaches also fail when the assumptions do not hold, which are often the cases in real-world images.  Moreover, most prior works, especially recent ones that utilize deep learning models, require a large amount of training data. That is, most of them are supervised learning methods, which produce acceptable results on images somewhat similar to the ones seen in the training datasets. Given the difficulty of collecting large (labeled) datasets for image reflection removal, these supervised learning methods would be likely to fail in many situations where the images have different characteristics than those in the training dataset.  

In this paper, we propose a novel \textit{unsupervised} method for single-image reflection separation, which does not require any additional information or large datasets to learn. 
Our method builds on the success of recent works that show that not all image priors must be learned from data. Rather, some of these characteristics can be captured by the network structure itself. This is referred to as Deep Image Prior (DIP) in the literature \cite{ulyanov2018deep}. 
DIP, however, is able to capture only low-level statistics of natural images. Thus, it may not produce good results for reflection separation, especially for natural images with reflection.  To address this problem, we design a new architecture of the network to contain high-level semantic information by embedding feature maps extracted from a pre-trained image classification network, and we refer to it as \textit{Perceptual DIP}. Also, our method composes two Perceptual DIPs with cross-feedback to generate both a background layer and a reflection layer, with good quality. 

The contributions of this paper can be summarized as follows.
\begin{itemize}
  \item We propose the first unsupervised method for single image reflection separation. Given only a single image observation, our method successfully generates background and reflection layers, without any training data or requiring additional information or assumptions. 
   
\item Our novel method is composed of two main parts: Perceptual DIP and cross-feedback. The first one is a new architecture of the generator network by embedding semantic features, allowing the network to utilize both low-level image statistics and high-level perceptual information during the optimization. The other encourages perceptually more meaningful separation by jointly optimizing the parameters of two Perceptual DIPs. 

%   \item We design a new architecture of the network by embedding semantic features, Perceptual DIP, which leads to more meaningful separation similar to human perception. 
%   \item We combine two Perceptual DIPs with cross-feedback in our method along with various optimization losses to improve the separation results. 
  
  \item We implement and compare our method versus five state-of-the-art methods, four of them are supervised learning methods \cite{percep, bdn, errnet, gcnet} and the fifth is unsupervised \cite{Gandelsman_2019_CVPR}. We utilize datasets commonly used in prior works and show that our method produces results that are at least as good as the ones produced by supervised learning methods, and on many occasions, much better results, especially on real-world images (i.e., not synthetic images). Our results show that our method consistently outperforms the closest unsupervised method, which also makes some assumptions about the inputs of the network while our method does not impose any restrictions. \end{itemize}

The rest of this paper is organized as follows. Section~\ref{sec:related} summarizes the related work in the literature. Section~\ref{sec:solution} presents the proposed solution. Section~\ref{sec:evaluation} compares the performance of the proposed method against the closest works in the literature, and Section~\ref{sec:conclusion} concludes the paper.

\section{Related Work} \label{sec:related}

% 1) Reflection Separation 
Since image reflection removal is a challenging task, it is necessary to exploit additional information to recover the underlying clean background. Some approaches use specialized devices or controlled capture settings to obtain a set of images of a target scene under different conditions such as varying focus \cite{schechner2000separation}, flash/no-flash  \cite{agrawal2005removing, agrawal2006edge}, multiple polarizer angles \cite{kong2013physically}, and recently two sub-aperture views from dual pixel sensors\cite{Punnappurath_2019_CVPR}. 

Many other approaches, however, have adopted post-processing methods using taken images or videos from ordinary users rather than skilled photographers. Especially when taking multiple images or videos from a slightly moving camera, we can observe the motion difference between background and reflection due to their different depths with respect to the camera (motion parallax). With this observation, a majority of the general multiple-image approaches are based on motion cue \cite{xue2015computational, sun2016automatic, Han_2017_CVPR, nandoriya2017video}, which significantly makes the problem more tractable. Recently, some works use a deep-learning framework \cite{alayrac2019visual, fan2019deep} to improve the performance. However, since multiple images from a scene are not always available in practice, the interest in single-image approaches has increased as they can easily access more extensive resources and extend to various applications.
\par %--deleted from motion cue ref (guo2014robust, li2013exploiting, gai2011blind, szeliski2000laye)r%

The single-image approaches leverage imposed assumptions or priors on reflection to make the problem more feasible. Some classical methods employ some heuristic prior knowledge from the observation, such as the sparse prior of gradients and local features  \cite{levin2004separating, levin2007user},  blurrier reflection prior \cite{li2014single}, the ghosting cues\cite{shih2015reflection} and different depth of fields between two layers \cite{wan2016depth}. Recently, the single-image reflection problem has shown notable achievements with deep-learning techniques \cite{fan2017generic, jin2018learning, percep, bdn, wan2018crrn, errnet, kim2019single}. While some earlier works use low-level losses on color and edges to train the networks \cite{fan2017generic, jin2018learning, wen2019single, lee2018generative}, Zhang et al. \cite{percep} improve the performance with perceptual losses by recognizing the high-level semantic meanings of the objects in different layers. Also, Yang et al. \cite{bdn} propose a cascade deep neural network (BDN) to estimates background and reflection bidirectionally. 

More recently, Abico et al. \cite{gcnet} introduce a gradient constraint loss along with the generative adversarial networks (GCNet) to produce high quality of the background layer. However, the supervised-learning techniques using the synthesized dataset reveal degraded performance on the real images. To tackle this problem, Wan et al. train the network on the aligned real dataset that they build \cite{wan2018crrn}, which is also released later for benchmarking other algorithms as the name of the single-image reflection dataset ($SIR^2$) \cite{sir}. Since acquiring aligned triplets of images($I,B,R$) is difficult in practice, Wei et al. \cite{errnet} propose the enhanced framework with context encoding module (ERRNet) to handle misaligned pairs of images. In a different way, some works attempt to generate more realistic synthesized datasets with physically-based rendering \cite{kim2019single}, non-linear blending formulation \cite{wen2019single} or generative adversarial training \cite{lee2018generative}. Nevertheless, none of the supervised methods above can fully overcome the limitation of degraded performance on images (especially natural scenes) that have not been seen in the training datasets. 

Newly, some works \cite{Gandelsman_2019_CVPR, chandramouli2019blind} attempt to tackle this problem in the limited manner of unsupervised approach with the help of the Deep Image Prior (DIP) \cite{ulyanov2018deep}. While most deep-learning techniques have focused on learning a realistic image prior over large datasets, the paper claims that a handcrafted structure of a generator network can be used as a deep image prior enough to capture low-level image statistics without any learning. This kind of approach is suitable for certain image restoration problems by optimizing the parameters of the untrained neural network to restore the target image from random noise. As an extension to this work, a unified framework using coupled deep-image-priors (Double-DIP) \cite{Gandelsman_2019_CVPR} is proposed for several unsupervised layer decomposition tasks including transparent layer separation. Based on the observation that the small patches of a natural image tend to have stronger internal self-similarity than the ones of a mixed image, a coupled DIP structure is enabled to separate the mixed image into its natural, simpler components. However, this approach only works well when the unknown latent layers in the single image are not correlated to each other or when they have two controlled images with different blending ratios using the same pair of layers, which is not applicable in natural setups. 

On the other hand, Chandramouli et al. \cite{chandramouli2019blind} exploit a generative model pretrained on facial images as a deep image prior to suppress unwanted reflections from a single face image. This method makes the problem less ill-posed by assuming the background layer as facial images. Thus this method can only handle face images and does not generalize to other types of images with reflection. 

In summary, our proposed method is the first unsupervised method for the single-image reflection separation problem, which does not require any training datasets or additional information.

\begin{figure*}[htp]
 \centering
 \includegraphics[width=1.0\textwidth]{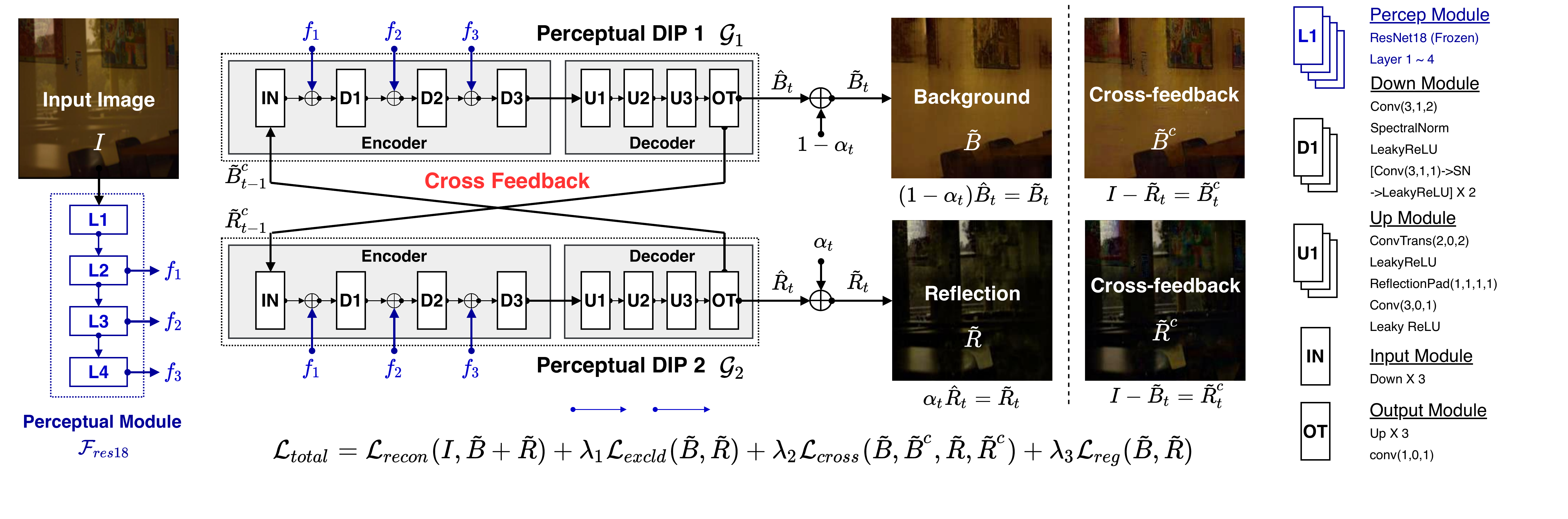}
 \caption{Overview of the unsupervised proposed method for single image reflection separation. Two Perceptual DIP networks with perceptual embedding are coupled with cross-feedback and loss functions, generating a background layer and a reflection layer from a given input image.
}
\label{framework}
\end{figure*}

\section{Proposed Method} \label{sec:solution}

This section describes the proposed unsupervised method for single image separation. 
Unlike the generic unsupervised layer decomposition method proposed in  \cite{Gandelsman_2019_CVPR} (Double DIP), our method aims explicitly to solve unsupervised reflection separation in natural images using uniquely designed perceptual DIPs. First, we introduce a novel design of the cross-coupled DIPs with perceptual embedding. Then, we describe the optimization algorithm with the corresponding loss functions. Figure \ref{framework} shows an overview of the proposed method, and the details are presented in the following subsections. 
 
\subsection{Architecture Design} 

\textbf{Perceptual Embedding:}
Employing perceptual cues has shown remarkable advantages in capturing semantic meanings for various image-related tasks. Several recent deep-learning techniques improve the performance with the combination of two perceptual losses: a feature loss to measure some distance in the high-level feature space from a pre-trained perceptual network, and an adversarial loss to generate realistic images by training a separate discriminator network in parallel. However, computing L1 or L2 distance between high-dimensional features is not sufficient to capture the real difference of them, plus an adversarial loss requires paired ground-truth datasets of background and reflection to discriminate real or fake data via supervised-learning.  
To overcome these weaknesses, we propose perceptual embedding, which contains multi-level feature maps directly fed to the corresponding layers of an encoder, rather than leveraging perceptual losses.
\par
\textbf{Perceptual DIP:}
Inspired by the perceptual discriminator \cite{sungatullina2018image}, we design an encoder-decoder style network with perceptual embedding, which is named as a \textit{Perceptual DIP}. At the initialization step, the perceptual embedding module extracts multi-level features from the pre-trained image classifier. As we choose ResNet18 \cite{he2016deep} as our backbone structure of the perceptual module, this module has four layers, but we skip the first layer output. This is because we observe that the features from this layer are more sensitive to low-level information of the image, similar to those captured by DIP, while our expectation for this module is to incorporate high-level features. Then, the extracted feature maps are concatenated with the features of each layer in the encoder, which is constructed to fit well with the size of the perceptual embedding and the input image. 

\begin{figure}[H]
 \centering
 \includegraphics[width=0.48\textwidth]{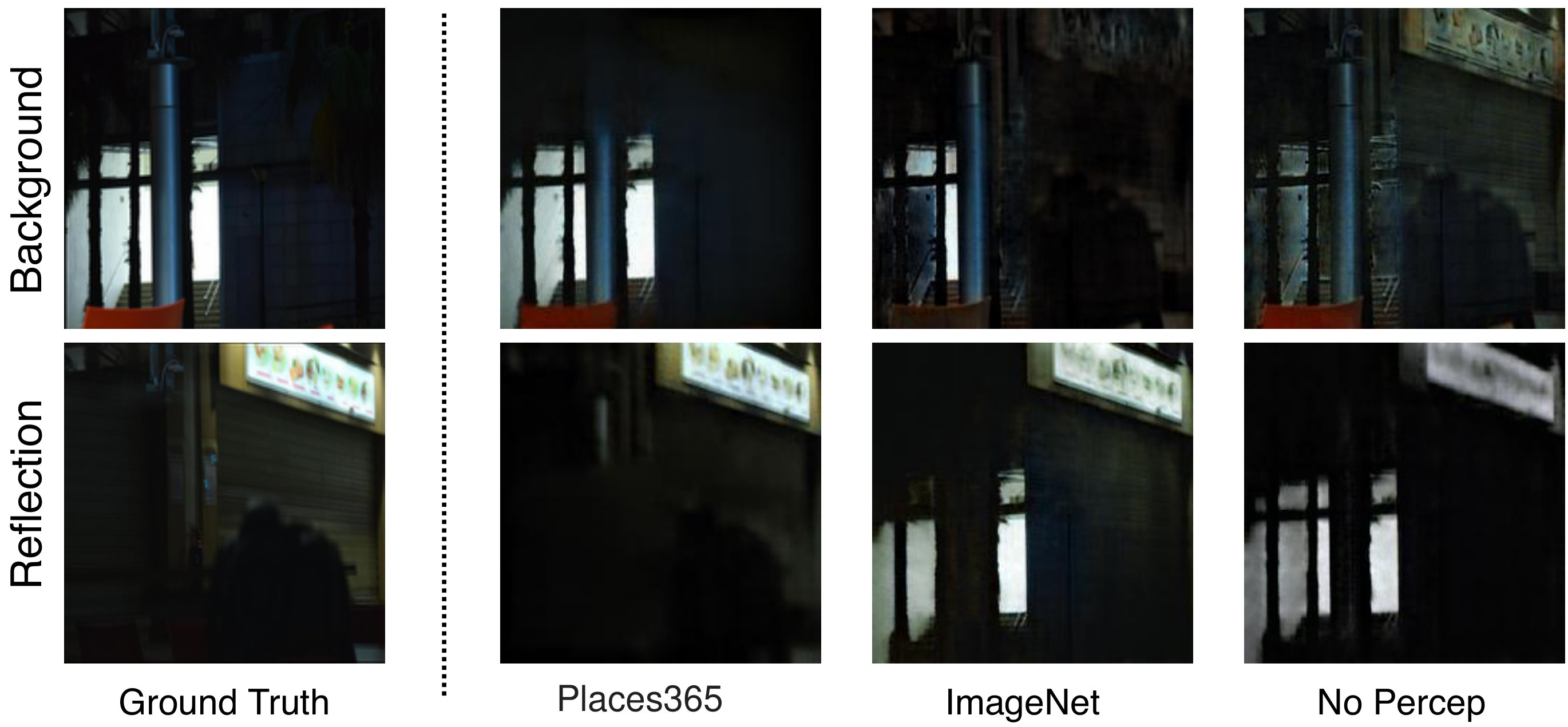}
 \caption{The impact of Perceptual Embedding on layer separation. Three different settings are shown: one using Places365~\cite{zhou2017places}, another using ImageNet~\cite{russakovsky2015imagenet}, and the third without Perceptual Embedding.}
\label{percep}
\end{figure}

The details of the corresponding down-sampling and up-sampling blocks in the network are shown in the Figure~\ref{framework}. We analyzed the impact of perceptual embedding on the reflection separation using various datasets. Sample results are shown in Figure~\ref{percep}, which indicates that the separation performance using the Places365 dataset \cite{zhou2017places} outperforms the one with the ImageNet dataset \cite{russakovsky2015imagenet} as well as the non-perceptual embedding case. 
We believe that it is because the Places365 dataset has more images for indoor and outdoor scenes, which usually exist in many reflection removal problems.

\par
\textbf{Cross-feedback:}
We use two coupled perceptual DIPs based on the observation that when two DIPs are jointly trained to reconstruct a single input image, each DIP tends to capture similar small patches inside the image while excluding each other \cite{Gandelsman_2019_CVPR}. While the original DIP \cite{ulyanov2018deep} generates the output from random noise, we feed the previous iteration output into the encoder to encourage the network to learn the difference between the given image and the updated feedback. Moreover, we upgrade the feedback into cross-feedback between two perceptual DIPs to enhance the ability of exclusion. Since one perceptual DIP outputs its estimation, we can automatically get a corresponding cross-estimation from Eq.~(\ref{eq1}) at each iteration $t$, i.e., $\tilde{B}_{t}^{c} = I-\tilde{R}_{t}$, and $\tilde{R}_{t}^{c} = I-\tilde{B}_{t}$, which is fed to the encoder of the other perceptual DIP. 

\begin{figure}[H]
 \centering
 \includegraphics[width=0.48\textwidth]{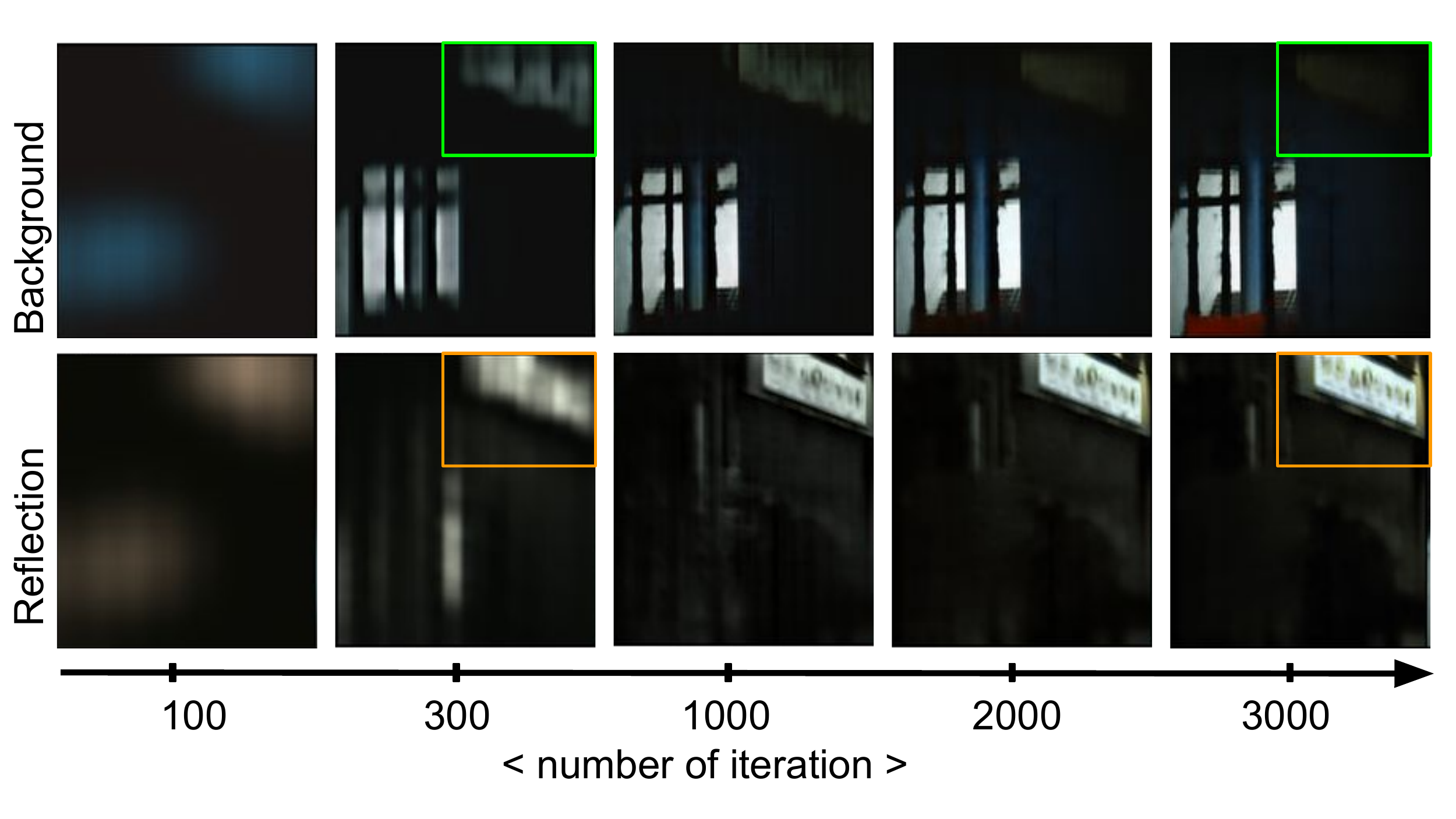}
 \caption{The effect of cross-feedback. At the early stage up to 300 iterations, both layers contain similar objects shown in the green and orange boxes. However, at iteration 3000, the reflection layer restores those objects while the background recovers other parts of the scene by excluding each other, similar to human perception.}
\label{cross}
\end{figure}

In Figure~\ref{cross}, we show how the two Perceptual DIPs are excluding each other throughout the iteration. From the observation, we find that the goal of the Perceptual DIP network is moved from restoration of the input image to exclusion between two layers as the number of iteration increases. 
\par

\subsection{Optimization Scheme}
\label{optim}
To consider Perceptual DIP in the optimization, we modify the technique introduced in the Deep Image Prior \cite{ulyanov2018deep}, which shows that the structure of the network is sufficient to capture a significant amount of image statistics without training on a large dataset. We define the structure of a Perceptual DIP as a parametric function $y = \mathcal{G}_{\theta}(x)$. Specifically, in our method, two Perceptual DIPs can be represented as $\hat{B}_{t} =  \mathcal{G}_{1}(\tilde{B}_{t-1}^{c},I)$ and  $\hat{R}_{t} =  \mathcal{G}_{2}(\tilde{R}_{t-1}^{c},I)$ given an input image $I$ and each cross-feedback, $\tilde{B}_{t-1}^{c} = I-\tilde{R}_{t-1}$ and $\tilde{R}_{t-1}^{c}=I-\tilde{B}_{t-1}$, at each iteration $t$. We note that the parameters of $\mathcal{G}_{\theta}$ do not include the ones of the fixed perceptual modules $\mathcal{F}_{res18}$.
  
 Besides, we add an external parameter $\alpha_{t}$ into the method to leverage which Perceptual DIP network generates which image layer based on the equation below. 
 \begin{align}
    \begin{cases}
        \tilde{B}_{t} &= (1-\alpha_{t}) \cdot \hat{B}_{t} \\
        \tilde{R}_{t} &= \alpha_{t} \cdot \hat{R}_{t}
    \end{cases},
  \qquad where \; \alpha_{t} \in (0, 0.5)
 \label{eq2}
\end{align}
where $\hat{B}_{t}$ and $\hat{R}_{t}$ are the direct outputs from two Perceptual DIP networks. The range of $\alpha$ is limited to under 0.5, as the range of (0.5, 1) would have the same effect. We set the initial guess of $ \alpha $ as 0.1, which implies that natural reflections are relatively weaker than the background scene in general cases. 
% \textcolor{red}
The impact of $\alpha$ is evaluated in the Section \ref{alpha}.
% }.

Based on this parameterization, we need to define the clear objectives of the optimization to find the perceptually meaningful decomposition of $\tilde{B}$ and $\tilde{R}$ from the input $I$. First, we list the following essential principles for layer separation: 
\begin{itemize}
\item The estimated outputs should be reconstructed based on the given image.  
\item The two recovered layers should be independent of one another. 
\item Each generated output should be a natural image.
\end{itemize}

Then, we realize these three principles with our loss functions: Reconstruction loss, Exclusive loss, and cross-feedback loss, and regularization loss, respectively. 
The total optimization loss can be written as:  
\begin{align}
     \mathcal{L}_{total} = \mathcal{L}_{recon} + \lambda_{1}\cdot \mathcal{L}_{excld} + \lambda_{2}\cdot \mathcal{L}_{cross} + \lambda_{3}\cdot \mathcal{L}_{reg},
\end{align}
where $\lambda_{i}$ is the corresponding weights for each loss functions based on the reconstruction loss. We experimentally measured the weights of different losses. Since the reconstruction loss performs the most important role in the problem definition, the other losses weight's were adjusted based on this loss to obtain better separation results. Once determined, we fixed them throughout the entire evaluation. Empirically, we set $\lambda_{1}$, $\lambda_{2}$ and $\lambda_{3}$ as 0.1, 0.1, and 1, respectively. The pseudo code of the optimization algorithm is shown in Algorithm~\ref{optimization} and the details of each loss are explained below. Also, our experiments on the impact of each loss are discussed in \ref{impact_loss}.

\par
\begin{algorithm}
    \caption{Optimization Algorithm}
    \raggedright
    \textbf{Require}: Decompose image $I$ into two latent layers: background $\tilde{B}$ and reflection $\tilde{R}$. 
    % \textcolor{red}
    {$ T$ denotes number of optimization iteration, which is fixed to 5000 through our experiments
    }
    \\
    \textbf{Input}: The image $I$ corrupted by unknown reflection \\
    \textbf{Output}: Decomposed layers, $\tilde{B}$ and $\tilde{R}$  \\
    \quad 1: initialize $\tilde{B}_{0} = \tilde{R}_{0}= I,  \alpha_{0} = 0.1$ \\ 
    \quad 2: \textbf{for} $t = 0$ to $T$: \\
    \quad 3: \qquad $\tilde{B}_{t} = (1-\alpha_{t}) \cdot \mathcal{G}_{1}(I - \tilde{R}_{t-1})$ \\
    \quad 4: \qquad $\tilde{R}_{t} = \alpha_{t} \cdot \mathcal{G}_{2}(I - \tilde{B}_{t-1})$ \\
    \quad 5: \qquad Compute the gradients of $\mathcal{L}_{total}$  \textit{w.r.t.}   $\tilde{B}_{t}, \tilde{R}_{t}, \alpha_{t} $ \\
    \quad 6: \qquad Update $\tilde{B}_{t}, \tilde{R}_{t}, \alpha_{t}$ using AdamW\cite{kingma2014adam}  \\
    \quad 7: \qquad $\tilde{B}_{t}^{c} = I-\tilde{R}_{t}$ \\ 
    \quad 8: \qquad $\tilde{R}_{t}^{c}=I-\tilde{B}_{t}$ \\
    \quad 9: \textbf{end for} \\
    \quad 10: $\tilde{B} = \tilde{B}_{T}, \tilde{R} = \tilde{R}_{T}$
    \label{optimization}
\end{algorithm}
\par

\textbf{Reconstruction Loss:}
We find that combining different types of reconstruction losses helps the network to converge faster. Thus, we define our reconstruction loss as: 
\begin{align}
    \mathcal{L}_{recon} &= \mathcal{L}_{color} + \omega_{1} \cdot \mathcal{L}_{gray} + \omega_{2} \cdot \mathcal{L}_{grad}, \\
    \mathcal{L}_{color} &= \|I - \tilde{I}\|_2, \nonumber\\
    \mathcal{L}_{gray}  &= \|    c(I) -     c(\tilde{I})\|_2, \nonumber\\
    \mathcal{L}_{grad}  &= \|\bigtriangledown_{x}I - \bigtriangledown_{x}\tilde{I}\|_1 +  \|\bigtriangledown_{y}I - \bigtriangledown_{y}\tilde{I}\|_1, \nonumber 
\end{align}
where $c(\cdot)$ is the conversion function from RGB image to gray-scale image, and $\bigtriangledown_{x,y}(\cdot)$ denotes the gradient of the input with the Sobel filter. The main reconstruction loss is a pixel-wise $\mathcal{L}2$ distance between the given image and the recombined image in RGB color space. We also design the same $\mathcal{L}2$ losses both in gray color space ($\mathcal{L}_{gray}$) and in gradient domain ($\mathcal{L}_{grad}$). We find that $\mathcal{L}_{gray}$ enhances the generated output and $\mathcal{L}_{grad}$ makes the network more robust. In the experiments, we set the value of $\omega_{1}$ and $\omega_{2}$ as 0.1. 
\par

\textbf{Exclusion Loss:}
The exclusion loss aims to minimize the correlation between two edges of the background and the reflection at multiple spatial resolutions, which enables us to reduce some residuals from each other. Thus similar to \cite{percep}, the exclusion loss is defined as:

% \textcolor{red}{added n}
\begin{align}
\label{eq5}
    \mathcal{L}_{excld} &= \sum_{n=1}^{N}\|norm(\bigtriangledown\tilde{B_{n}})\odot norm(\bigtriangledown\tilde{R_{n}})\|_{F},
\end{align}
% \textcolor{red}
where $n$ is the image down sampling factor, as exclusion loss minimizes the correlation between edges of background and reflection at multiple spatial resolution.  So each time in Eq.~(\ref{eq5}), the image is downsampled with a factor 2 and we chose N as 3 in the experiment. $norm(\cdot)$ is normalization in gradient fields of the two layers, $\odot$ is element-wise multiplication, and $\|\cdot\|_{F}$ denotes Frobenius norm.  

% \begin{figure}[ht!] %tp
%  \centering
%  \includegraphics[scale=1.0, width=0.9\linewidth]{graphics/loss_comp1.pdf}
%  \centering
%  \caption{Comparing the separation under two scenarios for losses: "I": Using only the Reconstruction Loss, "II": Reconstruction + Exclusion. The first and the second row are used to make a synthetic sample. }
% \label{loss_comp1}
% \end{figure}

\textbf{Cross-Feedback Loss:}
Our proposed design exploits cross-feedback to empower the network to exclude one another under the assumption that each generated layer should be similar to its corresponding cross-feedback from the other network as well as its previous output. We call the first constraint as the cross-consistent loss $\mathcal{L}_{cc}$ and the second one as the  feedback-consistent loss $\mathcal{L}_{fc}$, which are defined in the following: 
\begin{align}
    \mathcal{L}_{cross} &= \mathcal{L}_{cc} + \mathcal{L}_{fc}, \\
    {\mathcal{L}_{cc}}_{t} &= \|\tilde{B}_{t} - (I - \tilde{R}_{t-1})\|_{2} + \|\tilde{R}_{t} - (I - \tilde{B}_{t-1})\|_{2}, \nonumber\\
    {\mathcal{L}_{fc}}_{t} &= ssim(\tilde{B}_{t},\tilde{B}_{t-1}) + ssim(\tilde{R}_{t}, \tilde{R}_{t-1}). \nonumber
\end{align}
We find that using the $L2$ distance metric is better for cross-consistent loss, but for the feedback-consistent loss, the structural similarity metric ($ssim(\cdot)$) is more effective.

\textbf{Regularization Loss:}
We regulate the network under three priors: a total-variance loss $\mathcal{L}_{TV}$\cite{mahendran2015understanding},  a total-variance balance loss $\mathcal{L}_{TVB}$ that we applied on our own, and a ceiling rejection loss $\mathcal{L}_{ceil}$\cite{fan2019deep}, which are defined as follows:   
\begin{align}
    \mathcal{L}_{reg} &= \gamma_{1} \cdot \mathcal{L}_{TV} + \gamma_{2} \cdot \mathcal{L}_{TVB} + \mathcal{L}_{ceil}, \\
    \mathcal{L}_{TV}  &=  \|\bigtriangledown\tilde{B}_{t}\|_{1} + \|\bigtriangledown\tilde{R}_{t}\|_{1}, \nonumber\\
    \mathcal{L}_{TVB} &=  \|\bigtriangledown\tilde{B}_{t}\|_{1} - \|\bigtriangledown\tilde{R}_{t}\|_{1}, \nonumber\\
    \mathcal{L}_{ceil } &= \sum_{m} f(\tilde{B}_{t}, I, m) + f(\tilde{R}_{t}, I, m), \nonumber\\
    f(x,y,m) &= \begin{cases} \|x_{m} - y_{m}\|_{1} & if \; x_{m} > y_{m} \\ 0 & otherwise\end{cases},\nonumber
\end{align}

where $m$ denotes each image pixel. While a total-variance loss boosts the spatial smoothness in both generated scenes, our total-variance balance loss penalizes the system when one of the networks is giving up on generating the output (degeneration problem) by balancing the total gradients of each output. Also, ceiling rejection loss constrains each pixel whose intensity is larger than the input one, helping to resolve the color ambiguity. Empirically, $\gamma_{1}$ and $\gamma_{2}$ are set to 0.005 and 0.001, respectively.

% \begin{algorithm}
%     \caption{Optimization Algorithm}
%     \raggedright
%     \textbf{Require}: Decompose image $I$ into two latent layers: background $\tilde{B}$ and reflection $\tilde{R}$. 
%     \textcolor{red}{$ T$ denotes number of optimization iteration, which is fixed to 5000 through our experiments}
%     \\
%     \textbf{Input}: The image $I$ corrupted by unknown reflection \\
%     \textbf{Output}: Decomposed layers, $\tilde{B}$ and $\tilde{R}$  \\
%     \quad 1: initialize $\tilde{B}_{0} = \tilde{R}_{0}= I,  \alpha_{0} = 0.1$ \\ 
%     \quad 2: \textbf{for} $t = 0$ to $T$: \\
%     \quad 3: \qquad $\tilde{B}_{t} = (1-\alpha_{t}) \cdot \mathcal{G}_{1}(I - \tilde{R}_{t-1})$ \\
%     \quad 4: \qquad $\tilde{R}_{t} = \alpha_{t} \cdot \mathcal{G}_{2}(I - \tilde{B}_{t-1})$ \\
%     \quad 5: \qquad Compute the gradients of $\mathcal{L}_{total}$  \textit{w.r.t.}   $\tilde{B}_{t}, \tilde{R}_{t}, \alpha_{t} $ \\
%     \quad 6: \qquad Update $\tilde{B}_{t}, \tilde{R}_{t}, \alpha_{t}$ using AdamW\cite{kingma2014adam}  \\
%     \quad 7: \qquad $\tilde{B}_{t}^{c} = I-\tilde{R}_{t}$ \\ 
%     \quad 8: \qquad $\tilde{R}_{t}^{c}=I-\tilde{B}_{t}$ \\
%     \quad 9: \textbf{end for} \\
%     \quad 10: $\tilde{B} = \tilde{B}_{T}, \tilde{R} = \tilde{R}_{T}$
%     \label{optimization}
% \end{algorithm}
% \par
\begin{figure*}[ht!] %tp
 \centering
 \includegraphics[scale=1.5, width=0.96\textwidth]{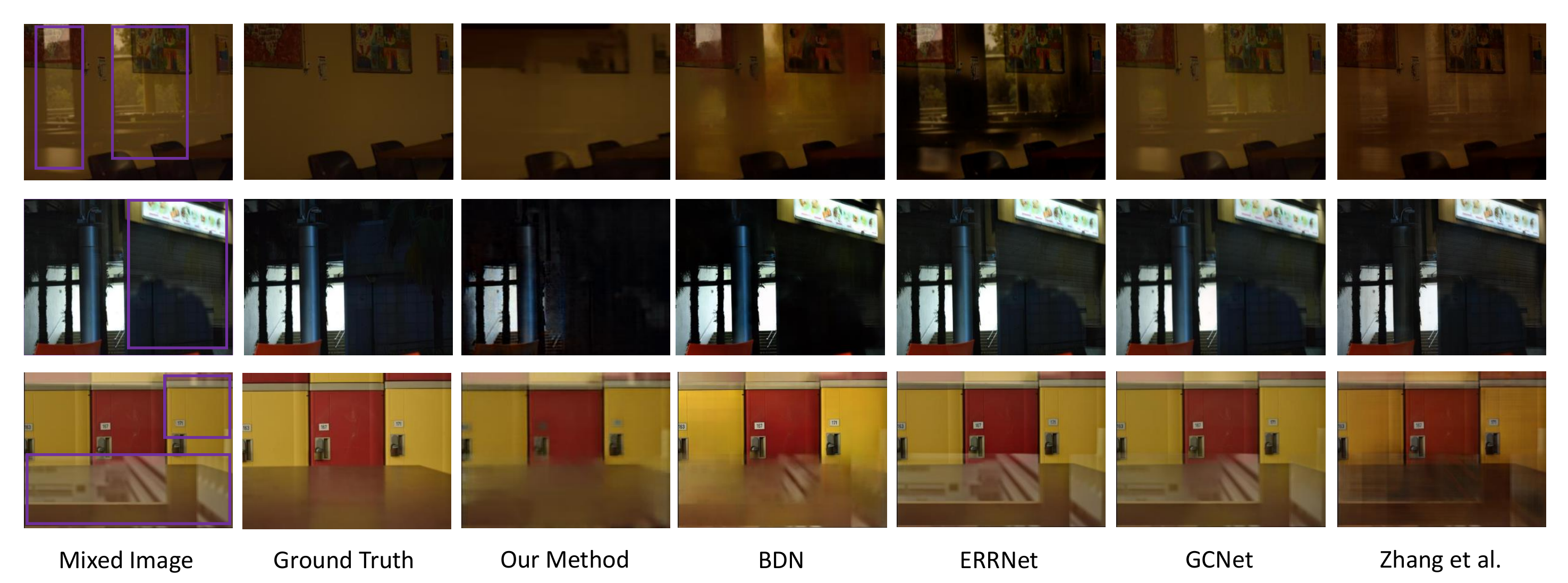}
 \centering
 \caption{Comparing our method versus four supervised methods on dataset DS1.}
\label{SIRcomp}
\end{figure*}

\begin{figure*}[h!] %tp
 \centering
 \includegraphics[scale=1.5, width=0.96\textwidth]{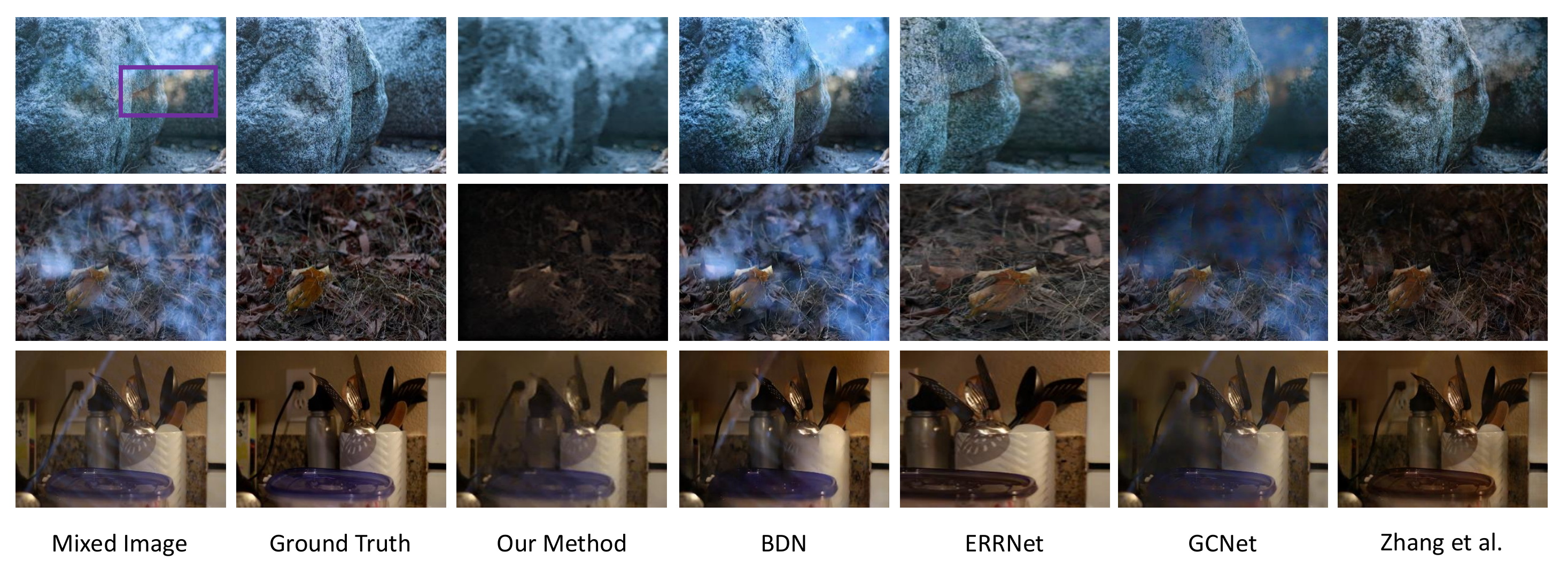}
 \centering
 \caption{Comparing our method versus four supervised methods on dataset DS2.}
\label{Bercomp}
\end{figure*}

\section{Evaluation} \label{sec:evaluation}

\subsection{Experimental Setup}

We evaluate the proposed reflection removal method and compare it against state-of-the-art methods using two real-world datasets that are commonly used to evaluate image reflection removal. The first dataset comes from \cite{sir}, and it contains 55 real-world images that contain reflection; we refer to it as DS1. These 55 images are the only real-world images with reflection having corresponding ground truth background and reflection layers in that dataset.
The second dataset, referred to as DS2, contains 20 images from the dataset in \cite{percep}. This dataset also has a ground truth background layer for each image. DS1 and DS2 contain diverse images of different indoor and outdoor scenes containing various levels of reflections.  
% \textcolor{red}

{Our datasets have about equal numbers of indoor and outdoor images. Figure~\ref{SIRcomp} has two rows of indoor scenes and two rows of outdoor scenes. And in Figure~\ref{Bercomp}, rows 1 and 2 contain outdoor scenes, while row 3 has indoor scenes.}

% \textcolor{red}
{Since our method is based on optimizing the model parameters on the single image input with the size of 224*224, the batch size is set as 1 and the parameters are updated with 0.0001 learning rate until the number of iteration(epochs) reaches to 5000}

We compare the proposed method against five state-of-the-art methods. Four of these method use supervised deep learning models, which are BDN \cite{bdn}, GCNet \cite{gcnet}, ERRNet \cite{errnet}, and Zhang et al. \cite{percep}. 

The fifth method we compare against is the unsupervised method proposed by Gandelsman et al.~\cite{chandramouli2019blind}, which we refer to as Double-DIP.  
Double-DIP takes two different mixtures of the same images as its input to accomplish the task of layer separation. As there was no specific method for mixing two layers mentioned in their paper to generate the second image, we experimented with two different settings. The first one we refer to as Double-DIP1, in which we mix the original background layer and the reflection layer that was modified by a Gaussian Kernel. As for the other setting, Double-DIP2, we linearly add two layers with a higher weight on the reflection layer to construct the second input.

For all of the five methods, we used the official implementations released by the authors of their papers.

We present sample images to show the qualitative comparison among the outputs of different methods. We also compare all methods quantitatively using the PSNR and SSIM metrics, as has been done in prior works in this area. \textbf{We note that the presented images are best viewed digitally and zoomed in to see the subtle differences}. We also note that we only present a few representative results due to space limitations. 

\begin{figure}[ht!] %tp
 \centering
 \includegraphics[scale=1.5, width=1.0\linewidth]{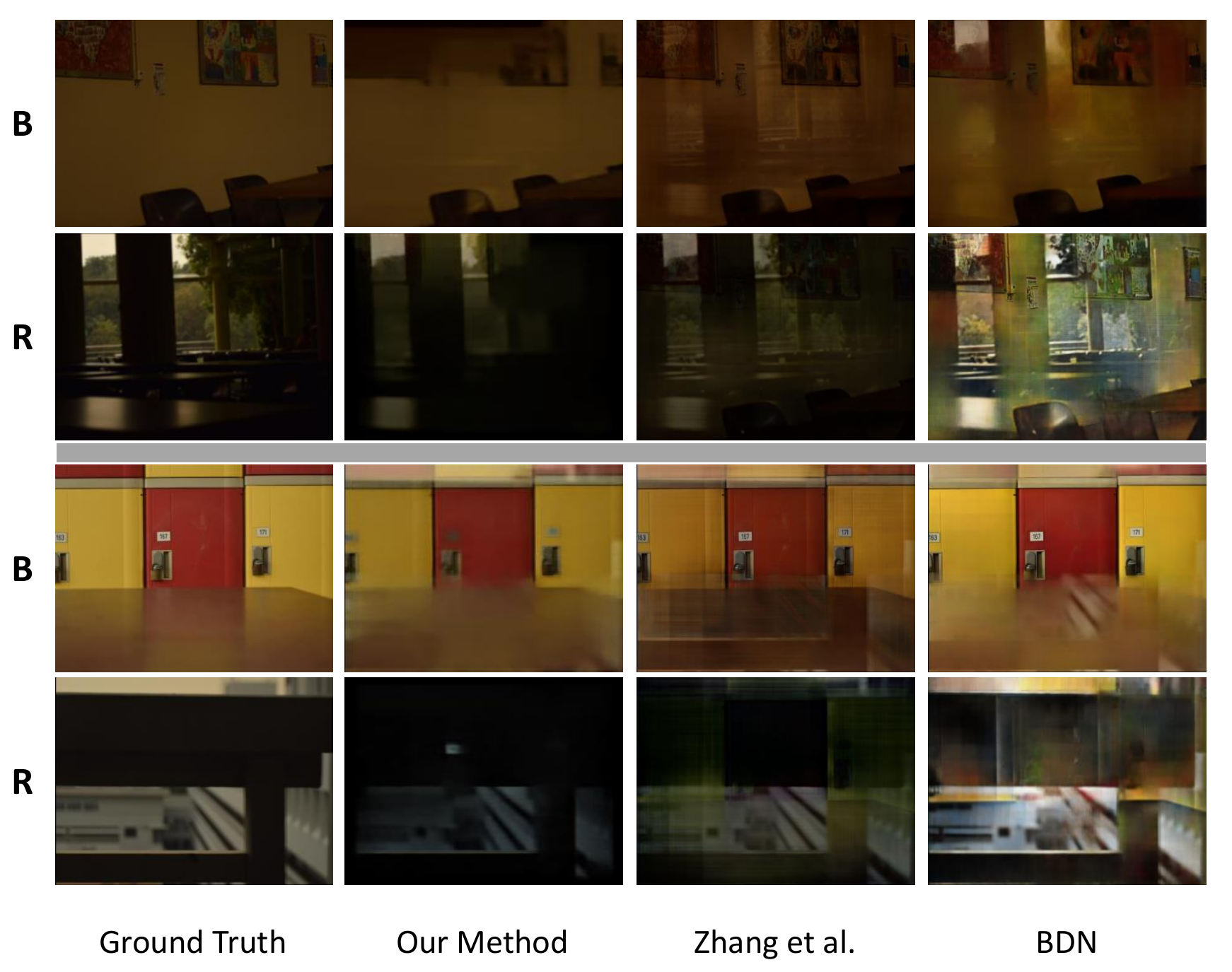}
 \centering
 \caption{Comparing the separation quality produced by our method versus BDN \cite{bdn} and Zhang et al. \cite{percep} methods. B and R indicate background and reflection, respectively}
\label{twolayercomp}
\end{figure}

\subsection{Comparison against Supervised Methods} 

We compare the proposed method versus four state-of-the-art supervised methods in Figure~\ref{SIRcomp} and Figure~\ref{Bercomp}, for datasets DS1 and DS2, respectively. 
In both figures, we draw rectangles showing some areas that have reflection. The input to all methods is shown on the left, which is an image with reflection. These two figures show only the background layer of each image after removing the reflection layer. We analyze the reflection layer later. 

The results in Figure~\ref{SIRcomp} and Figure~\ref{Bercomp} show that our method produces better (or at least the same) reflection removal than the supervised methods that require a substantial amount of training data. 
For example, in the sample image of the second row in Figure~\ref{SIRcomp}, all methods except ours failed to detect and remove the reflection.  Similarly, for the sample in the third row, our method generated an output close to the ground truth background, whereas the other models failed to remove the reflection in the image. The same observations can be made on the results in Figure~\ref{Bercomp}, which were produced on DS2. 

We further analyze the quality of the layer separation of different methods in Figure~\ref{twolayercomp}. This figure shows both the background and reflection layers produced by various methods and compares them against each other and the ground truth. We show only our method versus the BDN \cite{bdn} and  Zhang et al. \cite{percep} methods, as they were the ones that produced the best results from prior works, as indicated in Figure~\ref{SIRcomp} and \ref{Bercomp}. As Figure~\ref{twolayercomp} shows, our method produces a cleaner separation of the background and reflection layers. 

\begin{figure}[t!]
 \centering
 \includegraphics[scale=1.5, width=1\linewidth]{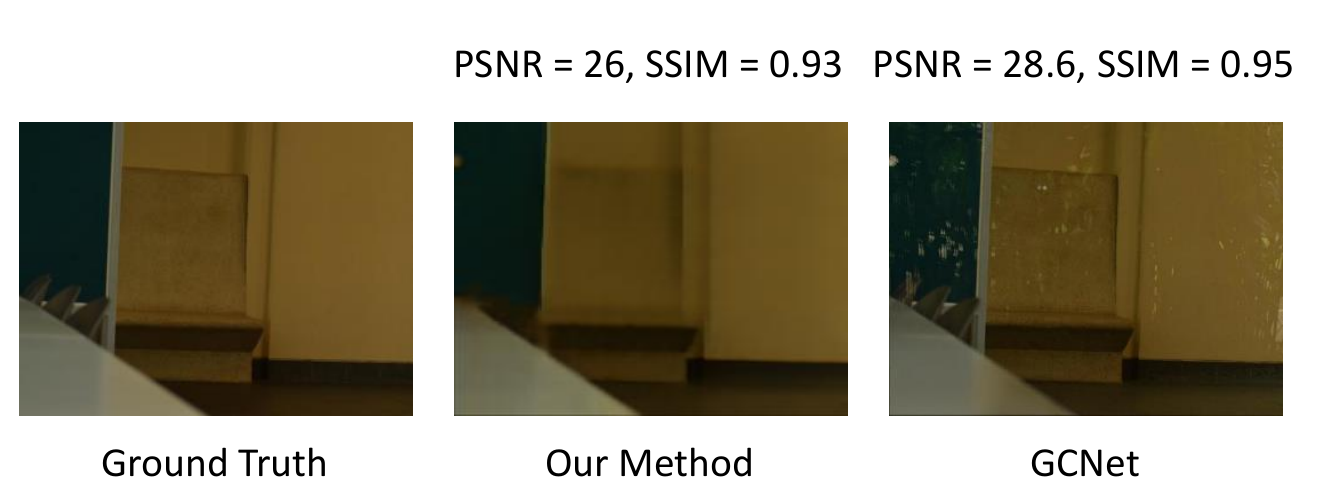}
 \centering
 \caption{Comparison between the output of our model and GCNet to show the importance of the visual quality over the objective PSNR and SSIM metrics. Although GCNet's output achieved better PSNR and SSIM, it did not remove much of the reflection, whereas our method removed most of the reflection.}
\label{PSNR}
\end{figure}

Next, we compare our method versus others using the PSNR and SSIM objective metrics, because such comparisons were made in previous works.   
The results for dataset DS1 are presented in Table~\ref{tbl:1}, which shows that our method results in somewhat smaller SSIM and PSNR values than some of the other methods. We note the SSIM and PSNR do not measure the quality of separation. Instead, they measure the quality of the produced images, even if the separation of the layer was not done properly. We illustrate this in Figure~\ref{PSNR}, where we compare the produced background layer of our method versus the one produced by GCNet. As the figure shows, GCNet produced a background that is similar to the input image without removing too much reflection. Thus, the computed PSNR and SSIM values are high, despite the poor performance in the main task at hand (removing reflection). On the other hand, our method removes most of the reflection from the image, but produces images with acceptable PSNR and SSIM values.

Supervised methods are data-driven, which means that they separate layers based on training their models with mostly synthetic datasets. We note that capturing natural images with reflection and creating ground-truth for them (i.e., the same scene but without reflection) is a very difficult task, especially for large datasets needed for deep learning models. 
This indicates that the performance of prior supervised methods heavily depends on the dataset and their performance typically degrades on images that do not have similar ones in the training dataset, which is usual for natural images. 
In contrast, our model is based on perceptual double-DIP, which exploits both high level and low-level statistics of an image to find two layers that are as close as possible to a natural image. And we optimize the parameters of the model on each input sample only, which means that it basically learns the image statistics of the input sample and uses them to separate the input into two layers.

It should be noted that reflection separation is a low-level vision task, but it is a very complex and ill-posed problem. To address this difficulty and reduce ambiguity, we utilize some high-level semantics. This is similar in nature to many prior works. 

We performed our evaluation on DS1 and DS2 dataset, which are also used in prior works. Our method mostly outperformed the supervised methods on DS1 samples, in which most images have a strong reflection.

\begin{table}[h!]
    %  \vspace{5} 
     \caption{Comparing our method versus other supervised learning methods using SSIM and PNSR metrics. B and R indicate background and reflection, respectively.}
    %  \caption{Dummy table}\label{tab:dummy-1}
     \begin{tabularx}{\linewidth}{c X}
     
    \vspace{-1pt}
    \makecell{ Dataset  \\   } &
    \hfill DS1  \hfill\null\\
  \midrule
  \vspace{2pt}
  \vspace{-1pt}
  \makecell{ Metric } & 
    \hfill \hspace{2pt}\makecell{PSNR \\ } 
    \hfill \hspace{15pt}\makecell{SSIM \\ } 
    \hfill\null \\
    \midrule

  \makecell{   } & 
    \hfill \hspace{0pt}\makecell{B \\ } 
    \hfill \hspace{8pt}\makecell{R \\ } 
    \hfill \hspace{6pt}\makecell{B  \\} 
    \hfill \hspace{3pt}\makecell{R \\} \hfill\null \\
    
    \midrule
    \vspace{5pt}
  \makecell{ BDN \cite{bdn} } & 
    \hfill \hspace{-1pt}\makecell{ 22.01 \\ } 
    \hfill \hspace{3pt}\makecell{9.01 \\ } 
    \hfill \hspace{1pt}\makecell{0.86  \\} 
    \hfill \makecell{0.31 \\} \hfill\null \\
    \vspace{5pt}

     \makecell{ GCNet \cite{gcnet} } & 
    \hfill \hspace{-7pt} \makecell{24.53 \\  }       
    \hfill \hspace{3pt}\makecell{ --- \\ } 
    \hfill \hspace{1pt}\makecell{ 0.92 \\} 
    \hfill \makecell{ --- \\} \hfill\null \\
    \vspace{5pt}
    
    \makecell{ Zhang et al. \cite{percep} } & 
    \hfill \makecell{21.13\\ } 
    \hfill \makecell{20.88 \\ } 
    \hfill \makecell{0.87  \\} 
    \hfill \makecell{0.64 \\} \hfill\null \\
    \vspace{5pt}
    
    \makecell{ ERRNet \cite{errnet} } & 
    \hfill \hspace{-6pt} \makecell{23.86 \\ } 
    \hfill \hspace{3pt}\makecell{ --- \\ } 
    \hfill \hspace{1pt}\makecell{0.88  \\} 
    \hfill \makecell{ ---\\} \hfill\null  \\
    \vspace{2pt}
    
    % \makecell{ Our Method 0.05 } & 
    % \hfill \makecell{ 20.95\\ } 
    % \hfill \makecell{ 20.02\\ } 
    % \hfill \makecell{0.81  \\} 
    % \hfill \makecell{0.61 \\} \hfill\null \\
    % \vspace{5}
    % \makecell{ Our Method 0.1 } & 
    % \hfill \hspace{1}\makecell{\textcolor{blue}{22.82} \\ } 
    % \hfill \hspace{1}\makecell{\textcolor{blue}{ \textbf{20.97} }  \\ } 
    % \hfill \hspace{-3}\makecell{\textcolor{blue}{0.83 } \\} 
    % \hfill \hspace{-2}\makecell{ \textbf{\textcolor{blue}{0.68}} \\} \hfill\null \\
    
    \makecell{ Our Method } & 
    \hfill \makecell{22.82\\ } 
    \hfill \makecell{20.97\\}    
    \hfill \makecell{0.83\\} 
    \hfill \makecell{0.68\\} \hfill\null \\
    
    % \vspace{2}
    % \makecell{ Our Method 0.5 } & 
    % \hfill \makecell{18.51 \\ } 
    % \hfill \makecell{18.44  \\ } 
    % \hfill \makecell{0.75  \\} 
    % \hfill \makecell{0.60 \\} \hfill\null \\
    
       \bottomrule
      
     \end{tabularx}
    %  \vspace{5} 
    %  \caption{Comparing our method versus other supervised learning methods using SSIM and PNSR metrics. B and R indicate background and reflection, respectively.}
\label{tbl:1}
   \end{table}

\subsection{Comparison against Unsupervised Method} 

We compare the proposed method against the closest unsupervised method in the literature \cite{Gandelsman_2019_CVPR}, which is referred to as Double-DIP. We note that we are aware of another unsupervised reflection removal work by Chandramouli et al. \cite{chandramouli2019blind}. However, this work focuses on removing reflection from face images, and it is not general like our method. Thus, we did not compare against it.

\begin{figure}[t!] %t
 \centering
 \includegraphics[scale=1.5, width=1.0\linewidth]{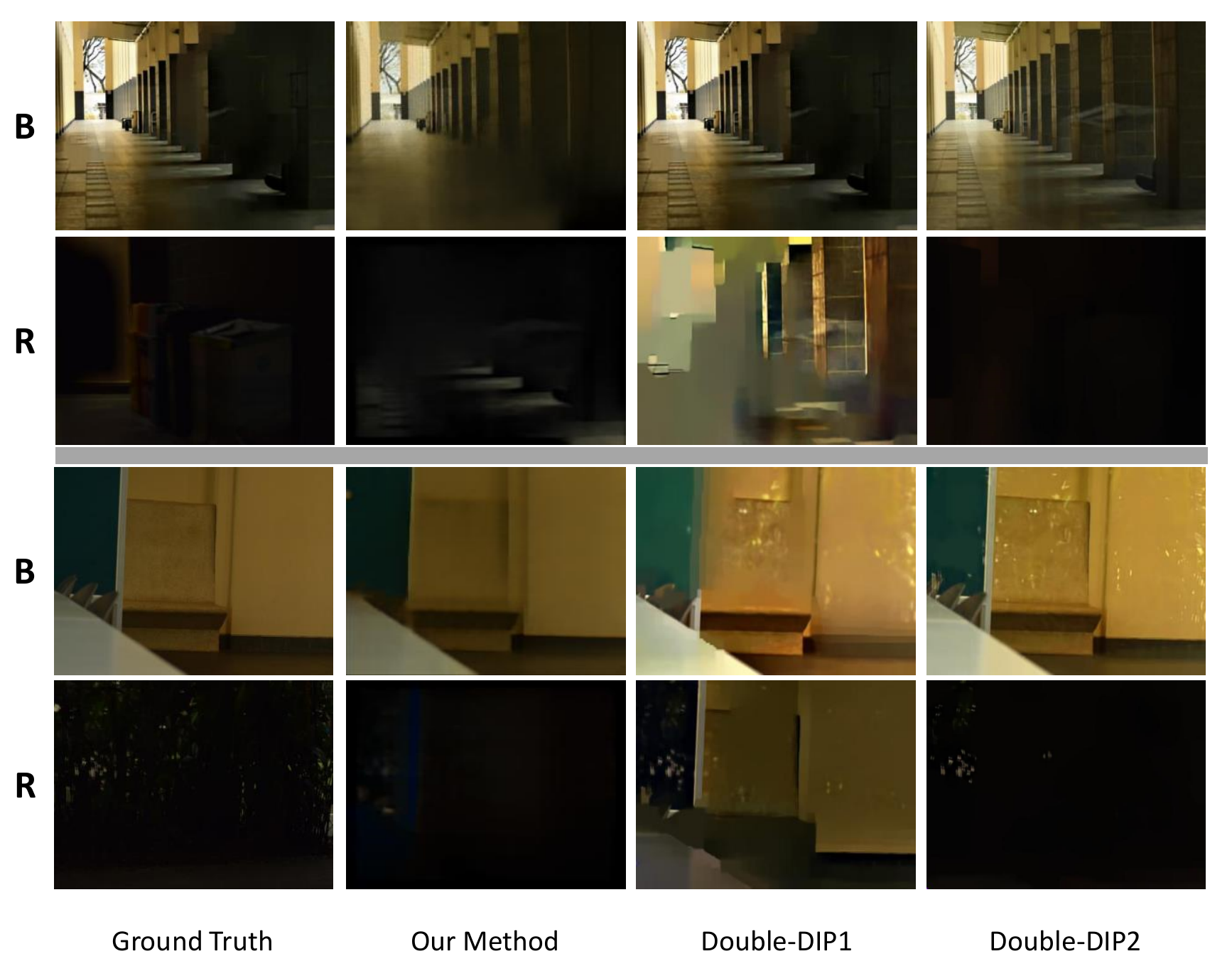}
 \centering
 \caption{Comparing our method against the unsupervised Double-DIP method  \cite{Gandelsman_2019_CVPR}. B and R indicate background and reflection, respectively. }
\label{unsup}
\end{figure}

Figure~\ref{unsup} compares our method versus Double-DIP using dataset DS1. 
As both reflection and background are needed for generating the second input image in Double-DIP, we could not evaluate it on the dataset DS2, which does not have the ground truth for the reflection layer. The results in the figure show that our method produces better results in terms of the separation quality. For example, as shown in the first two rows, our method performed better and separated the reflection from the background, whereas Double-DIP failed to remove the reflection. In the third row, Double-DIP tried to separate the mixed input into two different layers, but our method almost perfectly separated the reflection from the background. 

Next, we compare our method versus Double-DIP using PSNR and SSIM in Table~\ref{tbl:2} The table shows that our method achieves high PSNR and SSIM (especially for the background layer), while Double-DIP produces lower PSNR values. As commented before, PSNR and SSIM indicate the quality of the produced images, but they do not consider the layer separation quality.    

In summary, even though Double-DIP works as a transparent layer separator and takes two images as input, our method performs better in reflection separation both in the separation accuracy and the quality of the produced images.

\begin{table}[h!]
     \caption{Comparing our method versus another unsupervised learning method using SSIM and PNSR metrics. B and R indicate background and reflection, respectively.}
    %  \vspace{5} 
    %  \caption{Dummy table}\label{tab:dummy-1}
     \begin{tabularx}{\linewidth}{c X}
     
    \vspace{-1pt}
    \makecell{ Dataset  \\   } &
    \hfill DS1  \hfill\null\\
  \midrule
  \vspace{2pt}
  \vspace{-1pt}
  \makecell{ Metric } & 
    \hfill \hspace{2pt}\makecell{PSNR \\ } 
    \hfill \hspace{15pt}\makecell{SSIM \\ } 
    \hfill\null \\
    \midrule

  \makecell{   } & 
    \hfill \hspace{0pt}\makecell{B \\ } 
    \hfill \hspace{8pt}\makecell{R \\ } 
    \hfill \hspace{6pt}\makecell{B  \\} 
    \hfill \hspace{3pt}\makecell{R \\} \hfill\null \\
    
    \midrule
    \vspace{5pt}
  
    \makecell{ Double-DIP1 \cite{Gandelsman_2019_CVPR} } & 
    \hfill \makecell{ 16.61 \\ } 
    \hfill \makecell{10.02 \\ } 
    \hfill \makecell{0.73  \\} 
    \hfill \makecell{0.39 \\} \hfill\null \\
    \vspace{5pt}
    
    \makecell{ Double-DIP2 \cite{Gandelsman_2019_CVPR} } & 
    \hfill \makecell{16.53 \\ } 
    \hfill \makecell{ 20.35 \\ } 
    \hfill \makecell{  0.65 \\} 
    \hfill \makecell{0.66 \\} \hfill\null \\
    
    \vspace{2pt}
    \makecell{ Our Method } & 
    \hfill \makecell{22.82 \\ } 
    \hfill \makecell{20.97  \\ } 
    \hfill \makecell{0.83  \\} 
    \hfill \makecell{ 0.68 \\} \hfill\null \\

       \bottomrule
      
     \end{tabularx}
    %  \vspace{5} 
    %  \caption{Comparing our method versus another unsupervised learning method using SSIM and PNSR metrics. B and R indicate background and reflection, respectively.}
\label{tbl:2}
   \end{table}

\subsection{Impact of $\alpha$} 
\label{alpha}
\begin{figure}[h!] %t
 \centering
 \includegraphics[scale=1.5, width=1.0\linewidth]{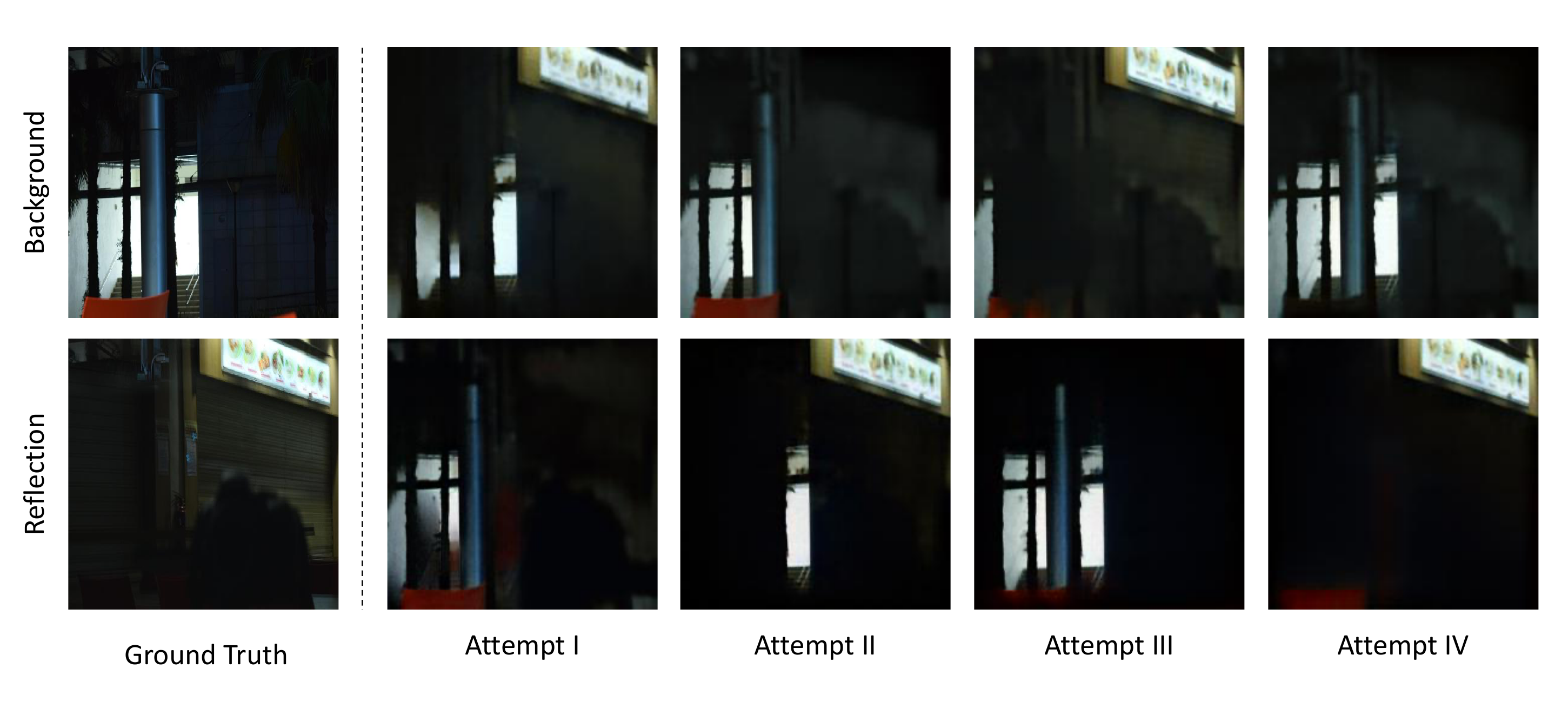}
 \caption{Inconsistent pairs of the outputs from multiple attempts on the same input when not using $\alpha$}
\label{alpha_inconsistant}
\end{figure}

\begin{figure}[h!] %t
 \centering
 \includegraphics[scale=1.5, width=1.0\linewidth]{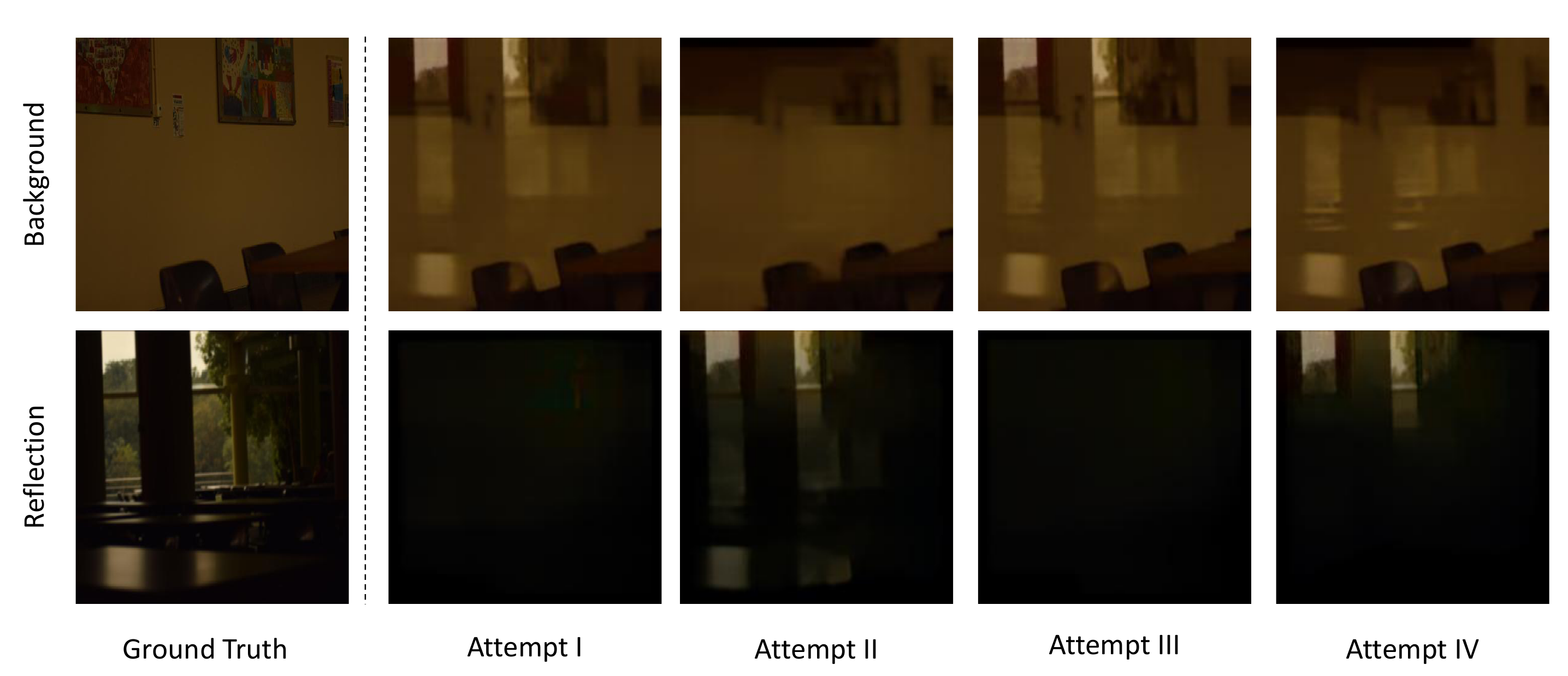}
 \centering
 \caption{Some degenerated pairs of the outputs (Attempt I and II) from multiple attempts when not using $\alpha$}
\label{alpha_degeneration}
\end{figure}

The image reflection removal problem is ill-posed, which means that there could be multiple solution pairs (background-reflection) for the same input image but what we need is only the desired solution. Without introducing the parameter $\alpha$, the problem definition Eq. (\ref{eq1}) is simple so that we could design our model with two identical Perceptual DIPs having an equal probability of generating each layer(background or reflection), which is not sufficient to resolve the ambiguity of the problem. This shortage of our model without $\alpha$ is observed through several failure cases in our repeated experiments. Figure \ref{alpha_inconsistant} shows that our model outputs inconsistent solution pairs when we test the model multiple times on the same input.  Also, for some samples with weak reflections, the model often gives up generating one of the layers as shown in Figure \ref{alpha_degeneration}. 

\begin{figure*}[h!t] %t
 \centering
 \includegraphics[scale=1.0, width=1.0\textwidth]{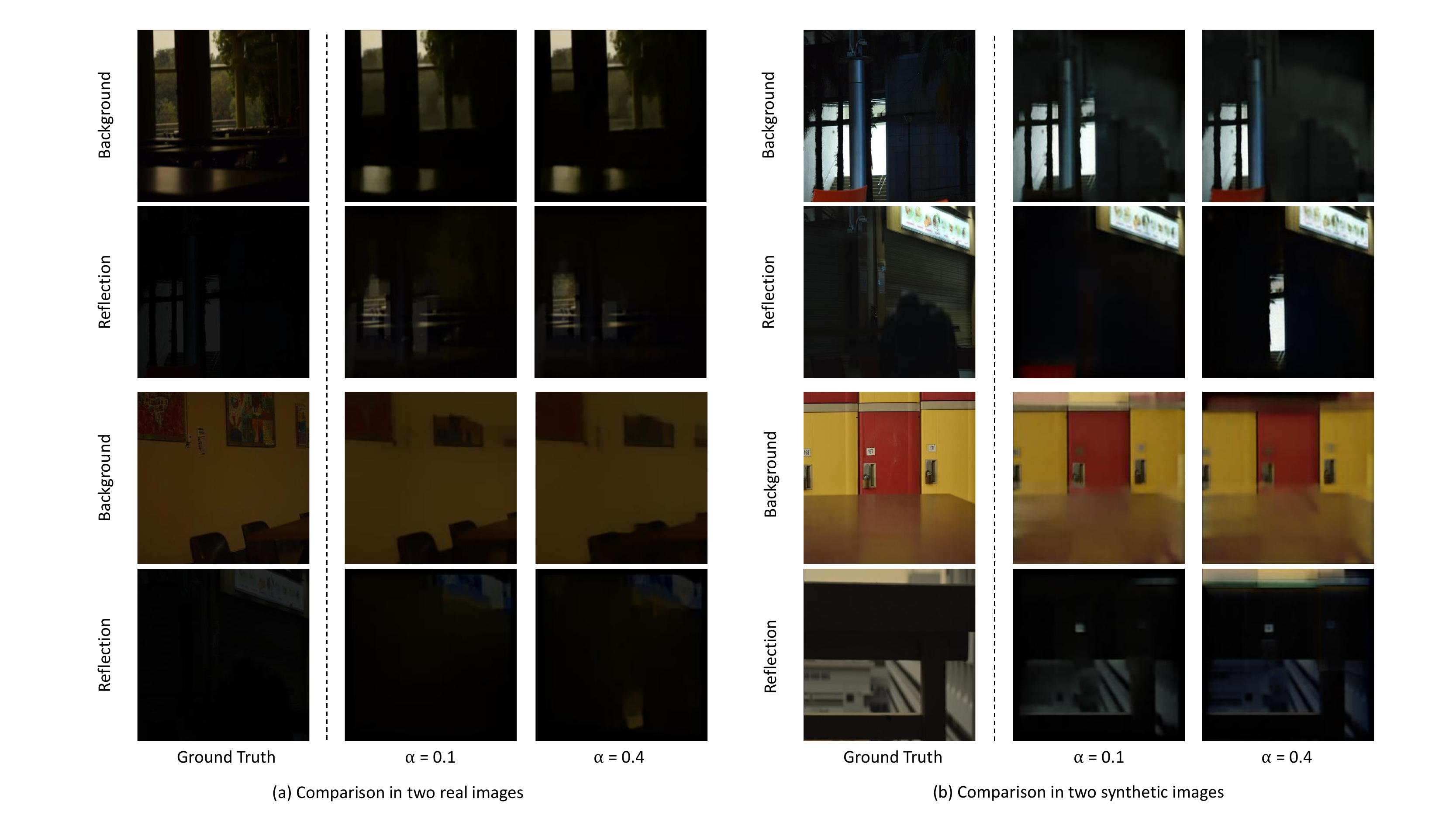}
 \centering
 \caption{Comparison between two $\alpha$ settings (0.1 and 0.4)}
\label{alpha_comparision}
\end{figure*}

 Parameter $\alpha$ mitigates this problem. Specifically, we observe that the reflection layer tends to have lower pixel intensity than the background layer in natural images. Thus, we model our problem in Eq. \ref{eq2} as a linear combination of two latent layers using $\alpha$, which assigns the role to each Perceptual DIP as either the background generator or reflection generator. In other words, we incorporate our prior belief of the balance between two unknown layers, and it gives us robust separation results from repeated experiments as well as significantly reduced degeneration issues on weak reflection samples. Through experimentation, we found that small $\alpha$ values (around 0.1) yield better results. As shown in Figure \ref{alpha_comparision}, the impact of $\alpha$ diminishes as we get closer to 0.5, as its influence on the two Perceptual DIPs becomes equal. Thus, we chose to use $\alpha$ = 0.1 for our final model and we do NOT change it. 

\subsection{Impact of different losses} 
\label{impact_loss}
\begin{figure*}[ht!]
 \centering
 \includegraphics[scale=1.0,width=0.7\textwidth]{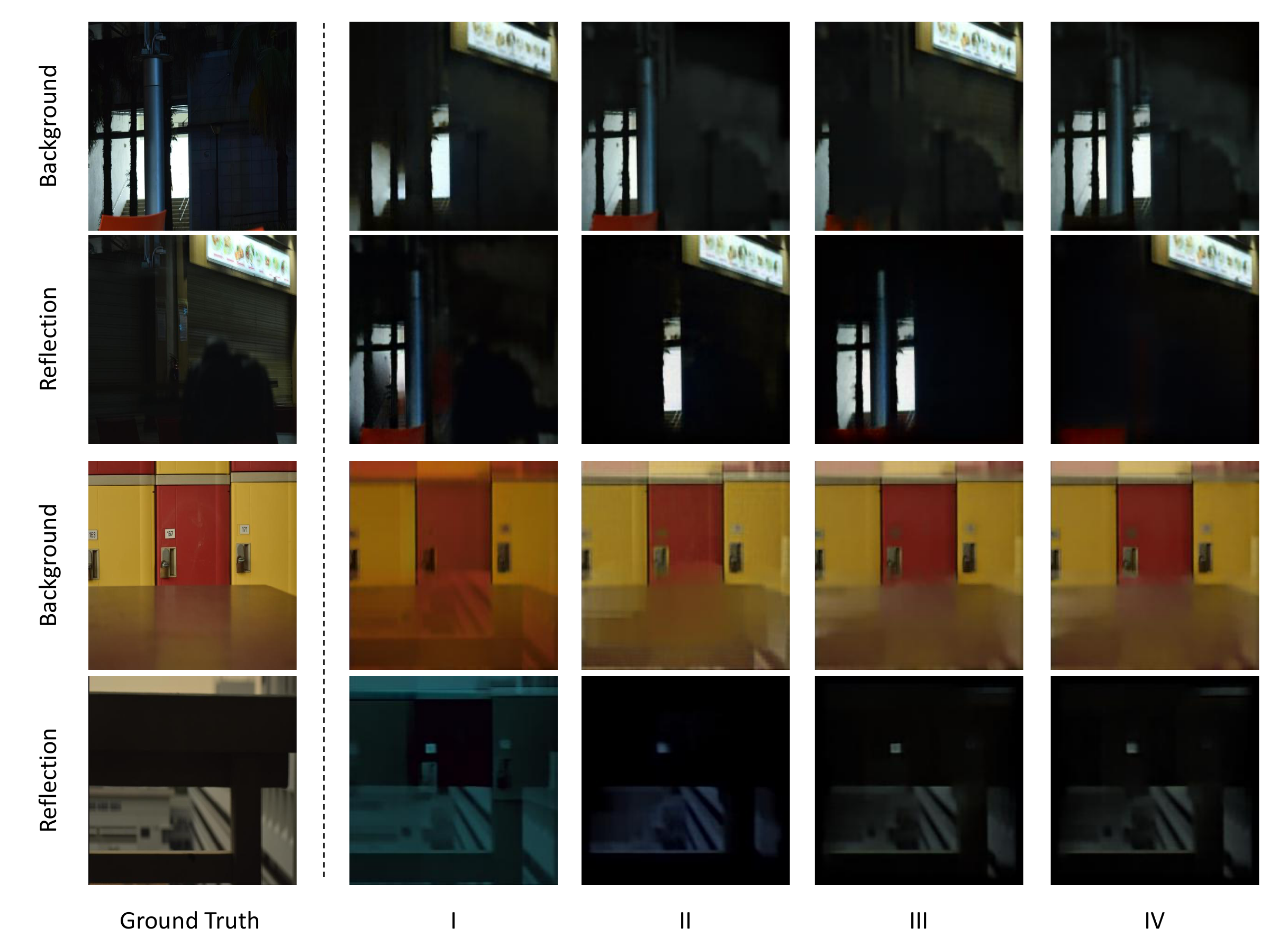}
 \caption{The impact of the proposed losses in four different scenarios in two real images:  "I": Using only the Reconstruction Loss, "II": Reconstruction + Exclusion, "III": Reconstruction + Exclusion + Cross-Feedback, and "IV": All the losses.}
\label{loss_study1}
\end{figure*}

\begin{figure*}[hb!]
 \centering
 \includegraphics[scale=1.0, width=0.7\textwidth]{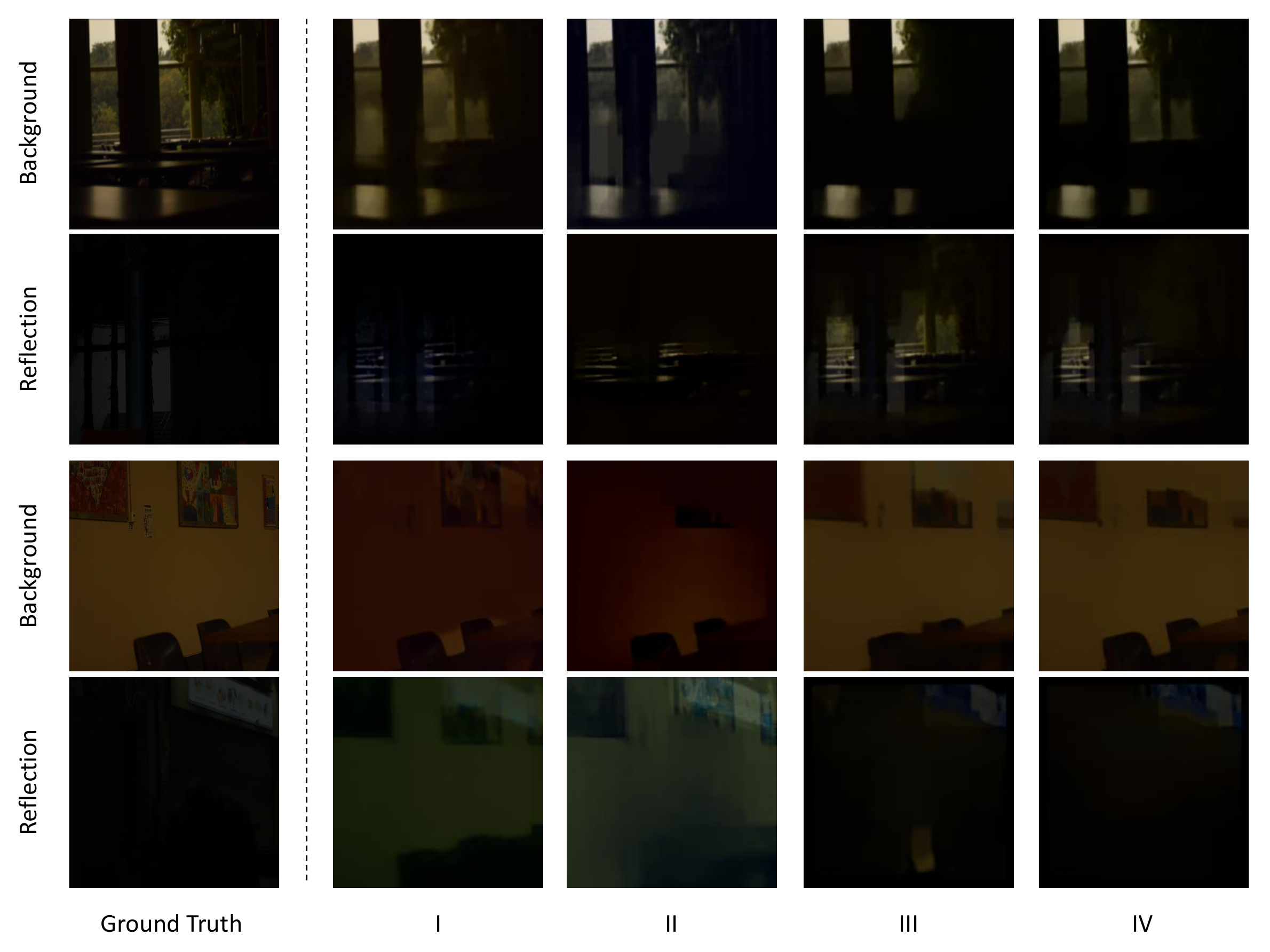}
 \caption{The impact of the proposed losses in four different scenarios in two synthetic image:  "I": Using only the Reconstruction Loss, "II": Reconstruction + Exclusion, "III": Reconstruction + Exclusion + Cross-Feedback, and "IV": All the losses.}
\label{loss_study2}
\end{figure*}

We design four types of loss terms as described in section \ref{optim}: reconstruction loss, exclusion loss, cross-feedback loss, and regularization loss. Since the reconstruction loss performs the most important role in the problem definition, we adjusted the weights of other losses based on this loss to obtain better separation results. Thus, we evaluate the impact of the different losses by adding each loss sequentially to the reconstruction loss as shown in Figure \ref{loss_study1} and Figure\ref{loss_study2} with a set of real and synthetic images. Since we utilize high-level features of Perceptual Embeddings, the separation result $I$, when using only reconstruction loss, looks reasonable but not sufficient due to high ambiguity between two latent layers. We append the exclusion loss to make the model decompose the input sample into two layers having different contents based on edge information, so $II$ shows better separation results than $I$ but still has some small artifacts. We enhance the exclusion with cross-feedback structure and its corresponding loss to perform well even when the gradient information of the reflection layer is not enough. While the result $III$ might be similar to $II$, the cross-feedback loss brings improvement in the speed of convergence and robustness of the model.  Finally, by joining the regularization term, we can obtain our best output shown in $IV$, which shows more solid separation in color and shapes.

% \textcolor{red}{
% In Figure \ref{loss_study}, We have addressed the impact of the four loss terms mainly. We start searching for weights from reconstruction loss. Having only the single reconstruction loss, the model is unstable showing some problems such as separation in color, degeneration of one of two layers, and inconsistent results from multiple attempts on the same sample (model robustness). We add the exclusion loss to make the model decompose the input sample into two layers having different contents based on edge information. Through experimentation, we find that the weight of this loss is 0.1 or less. However, the exclusion loss is not enough since it depends on the gradient of the input image. We design the cross-feedback structure and loss functions, which helps two Perceptual DIPs to exclude each other and use mutual information among themselves and brings improvement in speed of convergence and robustness, while the results might be similar. We find that 0.1 or less is good for the weight of this loss from the greed search. Finally, we solve the degeneration and low-robustness problem with the regularization terms. We observe the improvements in the colors and separates the image not only based on the shapes, but also in the contents. We set the weight for the ceiling of rejection loss as 1, 0.005 for total-variance loss, and 0.001 for total-variance-balance loss from the multiple experiments.}

\section{Conclusions and Future Work} \label{sec:conclusion}
We have presented an unsupervised method for single-image reflection removal. To the best of our knowledge, this is the first unsupervised work for removing reflection for natural scenes using only a single image. We have proposed a novel architecture of cross-coupled \textit{Perceptual DIPs} that is capable of capturing not only the low-level statistics of a natural image but also the high-level semantic cues. We have also designed an optimization scheme using multiple loss functions without training on any dataset, which significantly resolves the ambiguity of single-image separation and leads to good separation results for natural images. Both qualitative and quantitative evaluations on real datasets have shown that our method is on par with state-of-the-art supervised models or better in some cases, and significantly outperforms the closest unsupervised method in \cite{Gandelsman_2019_CVPR} that also needs to use an additional image, while our method does not. 

The work in this paper can be extended in multiple directions. For example, the quality of the two separated layers can further be improved by incorporating some suppression or inpainting techniques into our method to handle some extreme cases.

\balance
\clearpage

\bibliographystyle{ACM-Reference-Format}
\balance
\bibliography{main}


\begin{thebibliography}{36}

%%% the Web address.

\ifx \showCODEN    \undefined \def \showCODEN     #1{\unskip}     \fi
\ifx \showDOI      \undefined \def \showDOI       #1{#1}\fi
\ifx \showISBNx    \undefined \def \showISBNx     #1{\unskip}     \fi
\ifx \showISBNxiii \undefined \def \showISBNxiii  #1{\unskip}     \fi
\ifx \showISSN     \undefined \def \showISSN      #1{\unskip}     \fi
\ifx \showLCCN     \undefined \def \showLCCN      #1{\unskip}     \fi
\ifx \shownote     \undefined \def \shownote      #1{#1}          \fi
\ifx \showarticletitle \undefined \def \showarticletitle #1{#1}   \fi
\ifx \showURL      \undefined \def \showURL       {\relax}        \fi
% invisible to TeX
\providecommand\bibfield[2]{#2}
\providecommand\bibinfo[2]{#2}
\providecommand\natexlab[1]{#1}
\providecommand\showeprint[2][]{arXiv:#2}

\bibitem[\protect\citeauthoryear{Abiko and Ikehara}{Abiko and Ikehara}{2019}]%
        {gcnet}
\bibfield{author}{\bibinfo{person}{Ryo Abiko} {and} \bibinfo{person}{Masaaki
  Ikehara}.} \bibinfo{year}{2019}\natexlab{}.
\newblock \showarticletitle{Single Image Reflection Removal Based on GAN With
  Gradient Constraint}.
\newblock \bibinfo{journal}{\emph{IEEE Access}}  \bibinfo{volume}{7}
  (\bibinfo{year}{2019}), \bibinfo{pages}{148790--148799}.
\newblock


\bibitem[\protect\citeauthoryear{Agrawal, Raskar, and Chellappa}{Agrawal
  et~al\mbox{.}}{2006}]%
        {agrawal2006edge}
\bibfield{author}{\bibinfo{person}{Amit Agrawal}, \bibinfo{person}{Ramesh
  Raskar}, {and} \bibinfo{person}{Rama Chellappa}.}
  \bibinfo{year}{2006}\natexlab{}.
\newblock \showarticletitle{Edge suppression by gradient field transformation
  using cross-projection tensors}. In \bibinfo{booktitle}{\emph{2006 IEEE
  Computer Society Conference on Computer Vision and Pattern Recognition
  (CVPR'06)}}, Vol.~\bibinfo{volume}{2}. IEEE, \bibinfo{pages}{2301--2308}.
\newblock


\bibitem[\protect\citeauthoryear{Agrawal, Raskar, Nayar, and Li}{Agrawal
  et~al\mbox{.}}{2005}]%
        {agrawal2005removing}
\bibfield{author}{\bibinfo{person}{Amit Agrawal}, \bibinfo{person}{Ramesh
  Raskar}, \bibinfo{person}{Shree~K Nayar}, {and} \bibinfo{person}{Yuanzhen
  Li}.} \bibinfo{year}{2005}\natexlab{}.
\newblock \showarticletitle{Removing photography artifacts using gradient
  projection and flash-exposure sampling}.
\newblock In \bibinfo{booktitle}{\emph{ACM SIGGRAPH 2005 Papers}}.
  \bibinfo{pages}{828--835}.
\newblock


\bibitem[\protect\citeauthoryear{Alayrac, Carreira, and Zisserman}{Alayrac
  et~al\mbox{.}}{2019}]%
        {alayrac2019visual}
\bibfield{author}{\bibinfo{person}{Jean-Baptiste Alayrac},
  \bibinfo{person}{Joao Carreira}, {and} \bibinfo{person}{Andrew Zisserman}.}
  \bibinfo{year}{2019}\natexlab{}.
\newblock \showarticletitle{The visual centrifuge: Model-free layered video
  representations}. In \bibinfo{booktitle}{\emph{Proceedings of the IEEE
  Conference on Computer Vision and Pattern Recognition}}.
  \bibinfo{pages}{2457--2466}.
\newblock


\bibitem[\protect\citeauthoryear{Chandramouli and
  Vaishnavi~Gandikota}{Chandramouli and Vaishnavi~Gandikota}{2019}]%
        {chandramouli2019blind}
\bibfield{author}{\bibinfo{person}{Paramanand Chandramouli} {and}
  \bibinfo{person}{Kanchana Vaishnavi~Gandikota}.}
  \bibinfo{year}{2019}\natexlab{}.
\newblock \showarticletitle{Blind Single Image Reflection Suppression for Face
  Images using Deep Generative Priors}. In
  \bibinfo{booktitle}{\emph{Proceedings of the IEEE International Conference on
  Computer Vision Workshops}}. \bibinfo{pages}{0--0}.
\newblock


\bibitem[\protect\citeauthoryear{Fan, Yang, Hua, Chen, and Wipf}{Fan
  et~al\mbox{.}}{2017}]%
        {fan2017generic}
\bibfield{author}{\bibinfo{person}{Qingnan Fan}, \bibinfo{person}{Jiaolong
  Yang}, \bibinfo{person}{Gang Hua}, \bibinfo{person}{Baoquan Chen}, {and}
  \bibinfo{person}{David Wipf}.} \bibinfo{year}{2017}\natexlab{}.
\newblock \showarticletitle{A generic deep architecture for single image
  reflection removal and image smoothing}. In
  \bibinfo{booktitle}{\emph{Proceedings of the IEEE International Conference on
  Computer Vision}}. \bibinfo{pages}{3238--3247}.
\newblock


\bibitem[\protect\citeauthoryear{Fan, Yin, Chen, Wang, Aviles-Rivero, Li,
  Schnlieb, Lischinski, and Chen}{Fan et~al\mbox{.}}{2019}]%
        {fan2019deep}
\bibfield{author}{\bibinfo{person}{Qingnan Fan}, \bibinfo{person}{Yingda Yin},
  \bibinfo{person}{Dongdong Chen}, \bibinfo{person}{Yujie Wang},
  \bibinfo{person}{Angelica Aviles-Rivero}, \bibinfo{person}{Ruoteng Li},
  \bibinfo{person}{Carola-Bibiane Schnlieb}, \bibinfo{person}{Dani Lischinski},
  {and} \bibinfo{person}{Baoquan Chen}.} \bibinfo{year}{2019}\natexlab{}.
\newblock \showarticletitle{Deep Reflection Prior}.
\newblock \bibinfo{journal}{\emph{arXiv preprint arXiv:1912.03623}}
  (\bibinfo{year}{2019}).
\newblock


\bibitem[\protect\citeauthoryear{Gandelsman, Shocher, and Irani}{Gandelsman
  et~al\mbox{.}}{2019}]%
        {Gandelsman_2019_CVPR}
\bibfield{author}{\bibinfo{person}{Yosef Gandelsman}, \bibinfo{person}{Assaf
  Shocher}, {and} \bibinfo{person}{Michal Irani}.}
  \bibinfo{year}{2019}\natexlab{}.
\newblock \showarticletitle{"Double-DIP": Unsupervised Image Decomposition via
  Coupled Deep-Image-Priors}. In \bibinfo{booktitle}{\emph{The IEEE Conference
  on Computer Vision and Pattern Recognition (CVPR)}}.
\newblock


\bibitem[\protect\citeauthoryear{Han and Sim}{Han and Sim}{2017}]%
        {Han_2017_CVPR}
\bibfield{author}{\bibinfo{person}{Byeong-Ju Han} {and}
  \bibinfo{person}{Jae-Young Sim}.} \bibinfo{year}{2017}\natexlab{}.
\newblock \showarticletitle{Reflection Removal Using Low-Rank Matrix
  Completion}. In \bibinfo{booktitle}{\emph{The IEEE Conference on Computer
  Vision and Pattern Recognition (CVPR)}}.
\newblock


\bibitem[\protect\citeauthoryear{He, Zhang, Ren, and Sun}{He
  et~al\mbox{.}}{2016}]%
        {he2016deep}
\bibfield{author}{\bibinfo{person}{Kaiming He}, \bibinfo{person}{Xiangyu
  Zhang}, \bibinfo{person}{Shaoqing Ren}, {and} \bibinfo{person}{Jian Sun}.}
  \bibinfo{year}{2016}\natexlab{}.
\newblock \showarticletitle{Deep residual learning for image recognition}. In
  \bibinfo{booktitle}{\emph{Proceedings of the IEEE conference on computer
  vision and pattern recognition}}. \bibinfo{pages}{770--778}.
\newblock


\bibitem[\protect\citeauthoryear{Jin, S{\"u}sstrunk, and Favaro}{Jin
  et~al\mbox{.}}{2018}]%
        {jin2018learning}
\bibfield{author}{\bibinfo{person}{Meiguang Jin}, \bibinfo{person}{Sabine
  S{\"u}sstrunk}, {and} \bibinfo{person}{Paolo Favaro}.}
  \bibinfo{year}{2018}\natexlab{}.
\newblock \showarticletitle{Learning to see through reflections}. In
  \bibinfo{booktitle}{\emph{2018 IEEE International Conference on Computational
  Photography (ICCP)}}. IEEE, \bibinfo{pages}{1--12}.
\newblock


\bibitem[\protect\citeauthoryear{Kim, Huo, and Yoon}{Kim et~al\mbox{.}}{2019}]%
        {kim2019single}
\bibfield{author}{\bibinfo{person}{Soomin Kim}, \bibinfo{person}{Yuchi Huo},
  {and} \bibinfo{person}{Sung-Eui Yoon}.} \bibinfo{year}{2019}\natexlab{}.
\newblock \showarticletitle{Single Image Reflection Removal with
  Physically-based Rendering}.
\newblock \bibinfo{journal}{\emph{arXiv preprint arXiv:1904.11934}}
  (\bibinfo{year}{2019}).
\newblock


\bibitem[\protect\citeauthoryear{Kingma and Ba}{Kingma and Ba}{2014}]%
        {kingma2014adam}
\bibfield{author}{\bibinfo{person}{Diederik~P Kingma} {and}
  \bibinfo{person}{Jimmy Ba}.} \bibinfo{year}{2014}\natexlab{}.
\newblock \showarticletitle{Adam: A method for stochastic optimization}.
\newblock \bibinfo{journal}{\emph{arXiv preprint arXiv:1412.6980}}
  (\bibinfo{year}{2014}).
\newblock


\bibitem[\protect\citeauthoryear{Kong, Tai, and Shin}{Kong
  et~al\mbox{.}}{2013}]%
        {kong2013physically}
\bibfield{author}{\bibinfo{person}{Naejin Kong}, \bibinfo{person}{Yu-Wing Tai},
  {and} \bibinfo{person}{Joseph~S Shin}.} \bibinfo{year}{2013}\natexlab{}.
\newblock \showarticletitle{A physically-based approach to reflection
  separation: from physical modeling to constrained optimization}.
\newblock \bibinfo{journal}{\emph{IEEE transactions on pattern analysis and
  machine intelligence}} \bibinfo{volume}{36}, \bibinfo{number}{2}
  (\bibinfo{year}{2013}), \bibinfo{pages}{209--221}.
\newblock


\bibitem[\protect\citeauthoryear{Lee, Yang, and Oh}{Lee et~al\mbox{.}}{2018}]%
        {lee2018generative}
\bibfield{author}{\bibinfo{person}{Donghoon Lee}, \bibinfo{person}{Ming-Hsuan
  Yang}, {and} \bibinfo{person}{Songhwai Oh}.} \bibinfo{year}{2018}\natexlab{}.
\newblock \showarticletitle{Generative single image reflection separation}.
\newblock \bibinfo{journal}{\emph{arXiv preprint arXiv:1801.04102}}
  (\bibinfo{year}{2018}).
\newblock


\bibitem[\protect\citeauthoryear{Levin and Weiss}{Levin and Weiss}{2007}]%
        {levin2007user}
\bibfield{author}{\bibinfo{person}{Anat Levin} {and} \bibinfo{person}{Yair
  Weiss}.} \bibinfo{year}{2007}\natexlab{}.
\newblock \showarticletitle{User assisted separation of reflections from a
  single image using a sparsity prior}.
\newblock \bibinfo{journal}{\emph{IEEE Transactions on Pattern Analysis and
  Machine Intelligence}} \bibinfo{volume}{29}, \bibinfo{number}{9}
  (\bibinfo{year}{2007}), \bibinfo{pages}{1647--1654}.
\newblock


\bibitem[\protect\citeauthoryear{Levin, Zomet, and Weiss}{Levin
  et~al\mbox{.}}{2004}]%
        {levin2004separating}
\bibfield{author}{\bibinfo{person}{Anat Levin}, \bibinfo{person}{Assaf Zomet},
  {and} \bibinfo{person}{Yair Weiss}.} \bibinfo{year}{2004}\natexlab{}.
\newblock \showarticletitle{Separating reflections from a single image using
  local features}. In \bibinfo{booktitle}{\emph{Proceedings of the 2004 IEEE
  Computer Society Conference on Computer Vision and Pattern Recognition, 2004.
  CVPR 2004.}}, Vol.~\bibinfo{volume}{1}. IEEE, \bibinfo{pages}{I--I}.
\newblock


\bibitem[\protect\citeauthoryear{Li and Brown}{Li and Brown}{2014}]%
        {li2014single}
\bibfield{author}{\bibinfo{person}{Yu Li} {and} \bibinfo{person}{Michael~S
  Brown}.} \bibinfo{year}{2014}\natexlab{}.
\newblock \showarticletitle{Single image layer separation using relative
  smoothness}. In \bibinfo{booktitle}{\emph{Proceedings of the IEEE Conference
  on Computer Vision and Pattern Recognition}}. \bibinfo{pages}{2752--2759}.
\newblock


\bibitem[\protect\citeauthoryear{Mahendran and Vedaldi}{Mahendran and
  Vedaldi}{2015}]%
        {mahendran2015understanding}
\bibfield{author}{\bibinfo{person}{Aravindh Mahendran} {and}
  \bibinfo{person}{Andrea Vedaldi}.} \bibinfo{year}{2015}\natexlab{}.
\newblock \showarticletitle{Understanding deep image representations by
  inverting them}. In \bibinfo{booktitle}{\emph{Proceedings of the IEEE
  conference on computer vision and pattern recognition}}.
  \bibinfo{pages}{5188--5196}.
\newblock


\bibitem[\protect\citeauthoryear{Nandoriya, Elgharib, Kim, Hefeeda, and
  Matusik}{Nandoriya et~al\mbox{.}}{2017}]%
        {nandoriya2017video}
\bibfield{author}{\bibinfo{person}{Ajay Nandoriya}, \bibinfo{person}{Mohamed
  Elgharib}, \bibinfo{person}{Changil Kim}, \bibinfo{person}{Mohamed Hefeeda},
  {and} \bibinfo{person}{Wojciech Matusik}.} \bibinfo{year}{2017}\natexlab{}.
\newblock \showarticletitle{Video reflection removal through spatio-temporal
  optimization}. In \bibinfo{booktitle}{\emph{Proceedings of the IEEE
  International Conference on Computer Vision}}. \bibinfo{pages}{2411--2419}.
\newblock


\bibitem[\protect\citeauthoryear{Punnappurath and Brown}{Punnappurath and
  Brown}{2019}]%
        {Punnappurath_2019_CVPR}
\bibfield{author}{\bibinfo{person}{Abhijith Punnappurath} {and}
  \bibinfo{person}{Michael~S. Brown}.} \bibinfo{year}{2019}\natexlab{}.
\newblock \showarticletitle{Reflection Removal Using a Dual-Pixel Sensor}. In
  \bibinfo{booktitle}{\emph{The IEEE Conference on Computer Vision and Pattern
  Recognition (CVPR)}}.
\newblock


\bibitem[\protect\citeauthoryear{Russakovsky, Deng, Su, Krause, Satheesh, Ma,
  Huang, Karpathy, Khosla, Bernstein, et~al\mbox{.}}{Russakovsky
  et~al\mbox{.}}{2015}]%
        {russakovsky2015imagenet}
\bibfield{author}{\bibinfo{person}{Olga Russakovsky}, \bibinfo{person}{Jia
  Deng}, \bibinfo{person}{Hao Su}, \bibinfo{person}{Jonathan Krause},
  \bibinfo{person}{Sanjeev Satheesh}, \bibinfo{person}{Sean Ma},
  \bibinfo{person}{Zhiheng Huang}, \bibinfo{person}{Andrej Karpathy},
  \bibinfo{person}{Aditya Khosla}, \bibinfo{person}{Michael Bernstein},
  {et~al\mbox{.}}} \bibinfo{year}{2015}\natexlab{}.
\newblock \showarticletitle{Imagenet large scale visual recognition challenge}.
\newblock \bibinfo{journal}{\emph{International journal of computer vision}}
  \bibinfo{volume}{115}, \bibinfo{number}{3} (\bibinfo{year}{2015}),
  \bibinfo{pages}{211--252}.
\newblock


\bibitem[\protect\citeauthoryear{Schechner, Kiryati, and Basri}{Schechner
  et~al\mbox{.}}{2000}]%
        {schechner2000separation}
\bibfield{author}{\bibinfo{person}{Yoav~Y Schechner}, \bibinfo{person}{Nahum
  Kiryati}, {and} \bibinfo{person}{Ronen Basri}.}
  \bibinfo{year}{2000}\natexlab{}.
\newblock \showarticletitle{Separation of transparent layers using focus}.
\newblock \bibinfo{journal}{\emph{International Journal of Computer Vision}}
  \bibinfo{volume}{39}, \bibinfo{number}{1} (\bibinfo{year}{2000}),
  \bibinfo{pages}{25--39}.
\newblock


\bibitem[\protect\citeauthoryear{Shih, Krishnan, Durand, and Freeman}{Shih
  et~al\mbox{.}}{2015}]%
        {shih2015reflection}
\bibfield{author}{\bibinfo{person}{YiChang Shih}, \bibinfo{person}{Dilip
  Krishnan}, \bibinfo{person}{Fredo Durand}, {and} \bibinfo{person}{William~T
  Freeman}.} \bibinfo{year}{2015}\natexlab{}.
\newblock \showarticletitle{Reflection removal using ghosting cues}. In
  \bibinfo{booktitle}{\emph{Proceedings of the IEEE Conference on Computer
  Vision and Pattern Recognition}}. \bibinfo{pages}{3193--3201}.
\newblock


\bibitem[\protect\citeauthoryear{Sun, Liu, Yang, Zeng, Wang, and Liu}{Sun
  et~al\mbox{.}}{2016}]%
        {sun2016automatic}
\bibfield{author}{\bibinfo{person}{Chao Sun}, \bibinfo{person}{Shuaicheng Liu},
  \bibinfo{person}{Taotao Yang}, \bibinfo{person}{Bing Zeng},
  \bibinfo{person}{Zhengning Wang}, {and} \bibinfo{person}{Guanghui Liu}.}
  \bibinfo{year}{2016}\natexlab{}.
\newblock \showarticletitle{Automatic reflection removal using gradient
  intensity and motion cues}. In \bibinfo{booktitle}{\emph{Proceedings of the
  24th ACM international conference on Multimedia}}. \bibinfo{pages}{466--470}.
\newblock


\bibitem[\protect\citeauthoryear{Sungatullina, Zakharov, Ulyanov, and
  Lempitsky}{Sungatullina et~al\mbox{.}}{2018}]%
        {sungatullina2018image}
\bibfield{author}{\bibinfo{person}{Diana Sungatullina}, \bibinfo{person}{Egor
  Zakharov}, \bibinfo{person}{Dmitry Ulyanov}, {and} \bibinfo{person}{Victor
  Lempitsky}.} \bibinfo{year}{2018}\natexlab{}.
\newblock \showarticletitle{Image manipulation with perceptual discriminators}.
  In \bibinfo{booktitle}{\emph{Proceedings of the European Conference on
  Computer Vision (ECCV)}}. \bibinfo{pages}{579--595}.
\newblock


\bibitem[\protect\citeauthoryear{Ulyanov, Vedaldi, and Lempitsky}{Ulyanov
  et~al\mbox{.}}{2018}]%
        {ulyanov2018deep}
\bibfield{author}{\bibinfo{person}{Dmitry Ulyanov}, \bibinfo{person}{Andrea
  Vedaldi}, {and} \bibinfo{person}{Victor Lempitsky}.}
  \bibinfo{year}{2018}\natexlab{}.
\newblock \showarticletitle{Deep image prior}. In
  \bibinfo{booktitle}{\emph{Proceedings of the IEEE Conference on Computer
  Vision and Pattern Recognition}}. \bibinfo{pages}{9446--9454}.
\newblock


\bibitem[\protect\citeauthoryear{Wan, Shi, Duan, Tan, and Kot}{Wan
  et~al\mbox{.}}{2017}]%
        {sir}
\bibfield{author}{\bibinfo{person}{Renjie Wan}, \bibinfo{person}{Boxin Shi},
  \bibinfo{person}{Ling-Yu Duan}, \bibinfo{person}{Ah-Hwee Tan}, {and}
  \bibinfo{person}{Alex~C Kot}.} \bibinfo{year}{2017}\natexlab{}.
\newblock \showarticletitle{Benchmarking single-image reflection removal
  algorithms}. In \bibinfo{booktitle}{\emph{Proceedings of the IEEE
  International Conference on Computer Vision}}. \bibinfo{pages}{3922--3930}.
\newblock


\bibitem[\protect\citeauthoryear{Wan, Shi, Duan, Tan, and Kot}{Wan
  et~al\mbox{.}}{2018}]%
        {wan2018crrn}
\bibfield{author}{\bibinfo{person}{Renjie Wan}, \bibinfo{person}{Boxin Shi},
  \bibinfo{person}{Ling-Yu Duan}, \bibinfo{person}{Ah-Hwee Tan}, {and}
  \bibinfo{person}{Alex~C Kot}.} \bibinfo{year}{2018}\natexlab{}.
\newblock \showarticletitle{Crrn: Multi-scale guided concurrent reflection
  removal network}. In \bibinfo{booktitle}{\emph{Proceedings of the IEEE
  Conference on Computer Vision and Pattern Recognition}}.
  \bibinfo{pages}{4777--4785}.
\newblock


\bibitem[\protect\citeauthoryear{Wan, Shi, Hwee, and Kot}{Wan
  et~al\mbox{.}}{2016}]%
        {wan2016depth}
\bibfield{author}{\bibinfo{person}{Renjie Wan}, \bibinfo{person}{Boxin Shi},
  \bibinfo{person}{Tan~Ah Hwee}, {and} \bibinfo{person}{Alex~C Kot}.}
  \bibinfo{year}{2016}\natexlab{}.
\newblock \showarticletitle{Depth of field guided reflection removal}. In
  \bibinfo{booktitle}{\emph{2016 IEEE International Conference on Image
  Processing (ICIP)}}. IEEE, \bibinfo{pages}{21--25}.
\newblock


\bibitem[\protect\citeauthoryear{Wei, Yang, Fu, Wipf, and Huang}{Wei
  et~al\mbox{.}}{2019}]%
        {errnet}
\bibfield{author}{\bibinfo{person}{Kaixuan Wei}, \bibinfo{person}{Jiaolong
  Yang}, \bibinfo{person}{Ying Fu}, \bibinfo{person}{David Wipf}, {and}
  \bibinfo{person}{Hua Huang}.} \bibinfo{year}{2019}\natexlab{}.
\newblock \showarticletitle{Single image reflection removal exploiting
  misaligned training data and network enhancements}. In
  \bibinfo{booktitle}{\emph{Proceedings of the IEEE Conference on Computer
  Vision and Pattern Recognition}}. \bibinfo{pages}{8178--8187}.
\newblock


\bibitem[\protect\citeauthoryear{Wen, Tan, Qin, Liu, Han, and He}{Wen
  et~al\mbox{.}}{2019}]%
        {wen2019single}
\bibfield{author}{\bibinfo{person}{Qiang Wen}, \bibinfo{person}{Yinjie Tan},
  \bibinfo{person}{Jing Qin}, \bibinfo{person}{Wenxi Liu},
  \bibinfo{person}{Guoqiang Han}, {and} \bibinfo{person}{Shengfeng He}.}
  \bibinfo{year}{2019}\natexlab{}.
\newblock \showarticletitle{Single image reflection removal beyond linearity}.
  In \bibinfo{booktitle}{\emph{Proceedings of the IEEE Conference on Computer
  Vision and Pattern Recognition}}. \bibinfo{pages}{3771--3779}.
\newblock


\bibitem[\protect\citeauthoryear{Xue, Rubinstein, Liu, and Freeman}{Xue
  et~al\mbox{.}}{2015}]%
        {xue2015computational}
\bibfield{author}{\bibinfo{person}{Tianfan Xue}, \bibinfo{person}{Michael
  Rubinstein}, \bibinfo{person}{Ce Liu}, {and} \bibinfo{person}{William~T
  Freeman}.} \bibinfo{year}{2015}\natexlab{}.
\newblock \showarticletitle{A computational approach for obstruction-free
  photography}.
\newblock \bibinfo{journal}{\emph{ACM Transactions on Graphics (TOG)}}
  \bibinfo{volume}{34}, \bibinfo{number}{4} (\bibinfo{year}{2015}),
  \bibinfo{pages}{1--11}.
\newblock


\bibitem[\protect\citeauthoryear{Yang, Gong, Liu, and Shi}{Yang
  et~al\mbox{.}}{2018}]%
        {bdn}
\bibfield{author}{\bibinfo{person}{Jie Yang}, \bibinfo{person}{Dong Gong},
  \bibinfo{person}{Lingqiao Liu}, {and} \bibinfo{person}{Qinfeng Shi}.}
  \bibinfo{year}{2018}\natexlab{}.
\newblock \showarticletitle{Seeing deeply and bidirectionally: A deep learning
  approach for single image reflection removal}. In
  \bibinfo{booktitle}{\emph{Proceedings of the European Conference on Computer
  Vision (ECCV)}}. \bibinfo{pages}{654--669}.
\newblock


\bibitem[\protect\citeauthoryear{Zhang, Ng, and Chen}{Zhang
  et~al\mbox{.}}{2018}]%
        {percep}
\bibfield{author}{\bibinfo{person}{Xuaner Zhang}, \bibinfo{person}{Ren Ng},
  {and} \bibinfo{person}{Qifeng Chen}.} \bibinfo{year}{2018}\natexlab{}.
\newblock \showarticletitle{Single image reflection separation with perceptual
  losses}. In \bibinfo{booktitle}{\emph{Proceedings of the IEEE Conference on
  Computer Vision and Pattern Recognition}}. \bibinfo{pages}{4786--4794}.
\newblock


\bibitem[\protect\citeauthoryear{Zhou, Lapedriza, Khosla, Oliva, and
  Torralba}{Zhou et~al\mbox{.}}{2017}]%
        {zhou2017places}
\bibfield{author}{\bibinfo{person}{Bolei Zhou}, \bibinfo{person}{Agata
  Lapedriza}, \bibinfo{person}{Aditya Khosla}, \bibinfo{person}{Aude Oliva},
  {and} \bibinfo{person}{Antonio Torralba}.} \bibinfo{year}{2017}\natexlab{}.
\newblock \showarticletitle{Places: A 10 million Image Database for Scene
  Recognition}.
\newblock \bibinfo{journal}{\emph{IEEE Transactions on Pattern Analysis and
  Machine Intelligence}} (\bibinfo{year}{2017}).
\newblock


\end{thebibliography}

\end{document}